\documentclass{article} 
\usepackage[T1]{fontenc}
\usepackage{iclr2027_conference,times}

\usepackage{amsmath,amsfonts,bm}

\def\eqref#1{equation~\ref{#1}}

\def\1{\bm{1}}

\DeclareMathAlphabet{\mathsfit}{\encodingdefault}{\sfdefault}{m}{sl}
\SetMathAlphabet{\mathsfit}{bold}{\encodingdefault}{\sfdefault}{bx}{n}

\usepackage{hyperref}
\usepackage{url}
\usepackage{graphicx}
\usepackage{dsfont}
\usepackage{booktabs}
\usepackage{array}
\usepackage{tabularx}

\iclrfinalcopy
\title{Structured Stateful Relational Scripts with a High-Quality Dataset
for Multi-Shot Audio-Visual Video Generation}

\author{%
Chen Zhao$^{1,2}$%
\thanks{Equal contribution. This work was done during Chen Zhao's internship at Meituan.}
\quad
Lehan Shen$^{1,*}$
\quad
Qilong Huang$^{2}$
\quad
Hongyu Li$^{2}$
\\
\textbf{Zhuoliang Kang}$^{2}$
\quad
\textbf{Xiaoming Wei}$^{2}$
\quad
\textbf{Jian Yang}$^{1}$
\quad
\textbf{Ying Tai}$^{1}$%
\thanks{Corresponding author.}
\\
$^{1}$State Key Laboratory for Novel Software Technology,
Nanjing University, China
\\
$^{2}$Meituan, China
\\[0.5em]
\href{https://chenzhaonju.github.io/msav-script/}
{\texttt{https://chenzhaonju.github.io/msav-script/}}
}

\begin{document}

\maketitle

\begin{abstract}
Coherent multi-shot audio-visual video generation requires maintaining consistent identities while depicting evolving states and coordinating sounds across shot boundaries. Although dense captions describe individual shots in detail, they rarely specify which entities persist, how their states change, or which entities and actions produce sounds. To address this limitation, we propose Structured Stateful Relational Scripts (SSRS), a textual supervision representation that separates persistent entity and voice identities from shot-specific visual and auditory states. SSRS explicitly encodes sound-source bindings and temporal relations, while a global audio-visual event sequence connects local actions and sounds, including events that continue across shot boundaries. We further introduce MSAV-Script-100K, a large-scale dataset of curated multi-shot audio-visual videos paired with detailed SSRS annotations, enabling models to learn identity persistence, state transitions, and event continuity from structured supervision. Experiments demonstrate improvements in cross-shot consistency, adherence to specified state changes, and audio-visual alignment, highlighting the effectiveness of our SSRS and MSAV-Script-100K for coherent multi-shot audio-visual generation. Our proposed dataset is available at \href{https://huggingface.co/datasets/shenlehan/MSAV-Script-100K}{MSAV-Script-100K}.
\end{abstract}

\vspace{-0.3 cm}
\section{Introduction}
\label{sec:introduction}
\vspace{-0.1 cm}
Recent advances in text-to-video (T2V) generation have substantially improved visual fidelity, motion realism, and prompt adherence~\citep{hunyuanvideo,wan,moviegen}. The field is now moving beyond silent video synthesis toward \emph{joint audio-visual generation}, where speech, sound effects, ambience, music, and visual content are generated in a unified manner~\citep{seedance15pro,ltx2,ovi,mova}. This progression naturally leads to the more challenging setting of \emph{multi-shot audio-visual generation}, which requires not only high-quality content within each shot, but also coherent identities, evolving states, ongoing events, and audio-visual correspondence across shot boundaries.

Recent works have explored multi-shot video generation through shot-specific control, cinematic transitions, cross-shot grounding, and long-form narrative modeling~\citep{shotadapter,cinetrans,multishotmaster,holocine}. However, most focus primarily on the visual stream. Extending multi-shot generation to joint audio-video introduces additional challenges, including speaker identity, dialogue continuation, sound-source correspondence, and cross-modal temporal alignment. While proprietary models such as Seedance~2.0 already demonstrate strong multi-shot audio-video capabilities, open models such as LTX-2 still leave substantial room for improvement~\citep{seedance20,ltx2,msavbench}. This gap can be largely attributed to two key bottlenecks.
\textbf{ First, high-quality training data remain limited.} Although recent efforts have advanced the development of long-form cinematic video datasets~\citep{cinetrans,holocine, cinedance,huang2026unityshots}, compact, diverse, and high-quality native multi-shot audio-video clips with dense cross-shot and cross-modal supervision remain underexplored for foundation-model training.  \textbf{Second, existing textual representations fail to sufficiently disentangle the dependencies.} Dense captions may provide rich descriptive information, yet free-form prose often entangles persistent identities, transient states, actions, dialogue, and auditory events within a single context~\citep{hunyuan2026scriptavideo}. Consequently, the generator must implicitly infer what should persist, what should change, and whether events across different shots represent a continuous process or independent occurrences.

To address the representation bottleneck, we introduce \textbf{Structured Stateful Relational Scripts (SSRS)}, a structured textual supervision  for multi-shot audio-visual generation. SSRS separates persistent visual and voice identities from shot-specific visual and auditory states, while representing visual evolution, auditory evolution, and audio-visual relations in distinct fields. It further connects local events through a global audio-visual event sequence, explicitly modeling cross-shot continuity. This is important because a \emph{shot boundary is not necessarily an event boundary}: speech, ambience, or other auditory events may continue across visual cuts. SSRS therefore organizes supervision around a simple principle: \emph{what should persist, what should change, and what should connect}.

To support such structured supervision at scale, we introduce \textbf{MSAV-Script-100K}, a large-scale, high-quality dataset of compact multi-shot audio-visual clips designed for foundation-model training. Unlike long-form cinematic corpora, MSAV-Script-100K focuses on short but genuine multi-shot sequences with native synchronized audio and dense cross-shot interactions. Each clip is paired with comprehensive audio-visual descriptions and SSRS annotations. We further formulate dense-to-SSRS conversion as source-constrained reorganization, preserving the underlying information while changing only its structure. This enables controlled comparison between dense and structured supervision on the same videos and caption content.

Our contributions are twofold: (1) we propose \textbf{SSRS}, a structured supervision representation that explicitly organizes state evolution and audio-visual dependencies across shots; and (2) we introduce \textbf{MSAV-Script-100K}, a large-scale native multi-shot audio-visual dataset with SSRS annotations. 

\vspace{-0.2 cm}
\section{Related Work}
\label{sec:related_work}
\vspace{-0.1 cm}
\paragraph{Video and joint audio-visual generation.}
Video Diffusion Models~\citep{ho2022videodiffusion} extended image diffusion to temporal synthesis, while VideoCrafter2~\citep{chen2024videocrafter2} improved video quality by combining low-quality videos with high-quality synthetic images. Subsequent foundation models, including CogVideoX~\citep{yang2025cogvideox}, HunyuanVideo~\citep{hunyuanvideo}, Wan~\citep{wan},  and LongCat-Video~\citep{meituan2025longcatvideo}, advanced visual quality, text alignment, and scalable generation through transformer-based architectures and improved training pipelines. Beyond silent video, Movie Gen~\citep{moviegen} incorporates video-conditioned audio generation, whereas LTX-2~\citep{ltx2}, and Ovi~\citep{ovi}, MOVA~\citep{mova} jointly model audio and video. Proprietary systems such as Veo~3~\citep{collins2025veo3}, Sora~2~\citep{openai2025sora2}, and Seedance~2.0~\citep{seedance20} have also demonstrated synchronized audio-video synthesis. 
\vspace{-0.1 cm}
\paragraph{Multi-shot audio-visual generation.}
Multi-shot visual generation has been explored through hierarchical narrative synthesis in MovieDreamer~\citep{zhao2025moviedreamer}, shot-specific conditioning in ShotAdapter~\citep{shotadapter}, transition control in CineTrans~\citep{cinetrans}, cross-shot grounding in MultiShotMaster~\citep{multishotmaster}, and holistic scene modeling in HoloCine~\citep{holocine}. Extending to joint audio-video generation, CineDance~\citep{cinedance} introduces long-form cinematic data with hierarchical dual-modal annotations and shared character and scene anchors. UnityShots~\citep{huang2026unityshots} combines long- and short-term visual memory with audio-visual boundary gating and reference-speaker conditioning. MSAVBench~\citep{msavbench} provides complementary evaluation of multi-shot audio-video capabilities. Our work complements these efforts with compact native multi-shot training clips and SSRS supervision. 
\vspace{-0.1 cm}
\paragraph{Generation-oriented captioning.}
Any2Caption~\citep{wu2025any2caption} translates heterogeneous conditions into captions for controllable synthesis, while VC4VG~\citep{du2025vc4vg} studies caption design and evaluation for T2V training. For audio-visual perception, video-SALMONN~2~\citep{tang2025videosalmonn2} and Omni-Captioner~\citep{ma2026omnicaptioner} improve descriptive completeness and faithfulness through preference optimization and tool-assisted annotation, respectively. AVSCap~\citep{wang2026avscap} and TCA-captioner~\citep{zhao2026temporal} emphasize cross-modal event binding. TimeChat-Captioner~\citep{yao2026timechatcaptioner} introduces time-aware, structured descriptions of multi-scene videos. Most closely related, Script-a-Video~\citep{hunyuan2026scriptavideo} proposes MTSS, factorizing descriptions into information streams connected by identity and temporal links. 
Building on prior work, SSRS combines global audio-visual context with shared registries for visual entities, voice identities, and audio entities. It distinguishes persistent identities from shot-specific states and voice identities from auditory events, separates shot-local visual evolution, auditory evolution, and cross-modal relations, and connects events across shots through a global audio-visual event sequence. 

\begin{figure}[t]
    \centering
    \includegraphics[width=\linewidth]{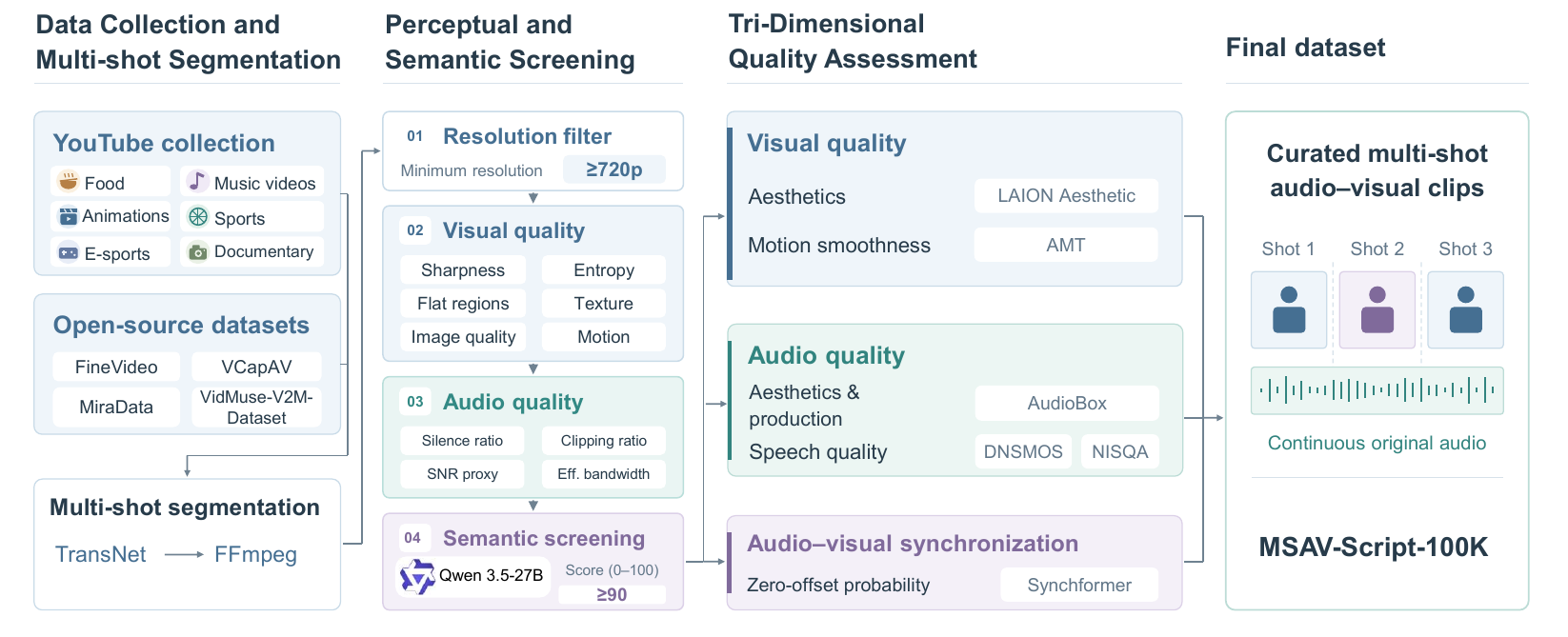}
    \vspace{-0.3 cm}
\caption{\textbf{Overview of the MSAV-Script-100K
construction pipeline.}
Videos collected from YouTube and existing datasets
are segmented into multi-shot clips, followed by
perceptual and semantic screening and tri-dimensional
assessment of visual quality, audio quality, and
audio-visual synchronization, yielding a  dataset
that preserves consecutive shots and continuous source audio. 
}
    \vspace{-0.2 cm}
    \label{fig:msav_script_pipeline}
\end{figure}

\section{MSAV-Script-100K}

\subsection{Data Collection and Multi-Shot Segmentation}


\paragraph{Data Collection.}

Constructing high-quality data for multi-shot audio-visual
generation requires native audio, coherent shot sequences, and
diverse cross-shot interactions. Existing datasets provide useful sources, but are often designed
for other tasks or dominated by cinematic content, with additional
limitations in resolution and visual and audio quality that restrict
their coverage of diverse, high-quality real-world audio-visual
scenarios.
 We therefore combine videos from existing
open datasets, including
FineVideo~\citep{Farre2024FineVideo},
MiraData~\citep{ju2024miradatalargescalevideodataset},
VCapAV~\citep{wang2025vcapav}, and
VidMuse-V2M-Dataset~\citep{tian2025vidmusesimplevideotomusicgeneration},
with additional high-resolution videos collected from YouTube.
For VCapAV~\citep{wang2025vcapav}, we use only real videos paired with their original audio tracks, excluding all samples containing synthetic video or audio.

Our YouTube collection covers six categories:
\emph{food, music videos, animations, documentaries, sports,
and esports}. These sources broaden the range of scenes,
subjects, and audio-visual events, including cooking, musical
performance, dialogue, narration, athletic actions, and commentated gameplay. They provide examples of recurring entities, evolving actions, and auditory events that persist across visual cuts, forming the source pool for multi-shot segmentation.


\paragraph{Multi-Shot Segmentation.}
We use TransNet V2~\citep{soucek2020transnetv2effectivedeep} to
detect shot boundaries and FFmpeg to split the source videos.
We then group consecutive shots into multi-shot clips with durations of approximately 6--12 seconds, preserving their original temporal order and corresponding audio. This retains natural
shot transitions and cross-cut audio continuity, providing compact
sequences for modeling cross-shot audio-visual dependencies.

This collection and segmentation process yields a pool of approximately 240K multi-shot audio-visual clips, referred to as the \textit{raw data}, which undergo multi-stage quality curation.


\begin{figure}[t]
    \centering
    \includegraphics[width=\linewidth]{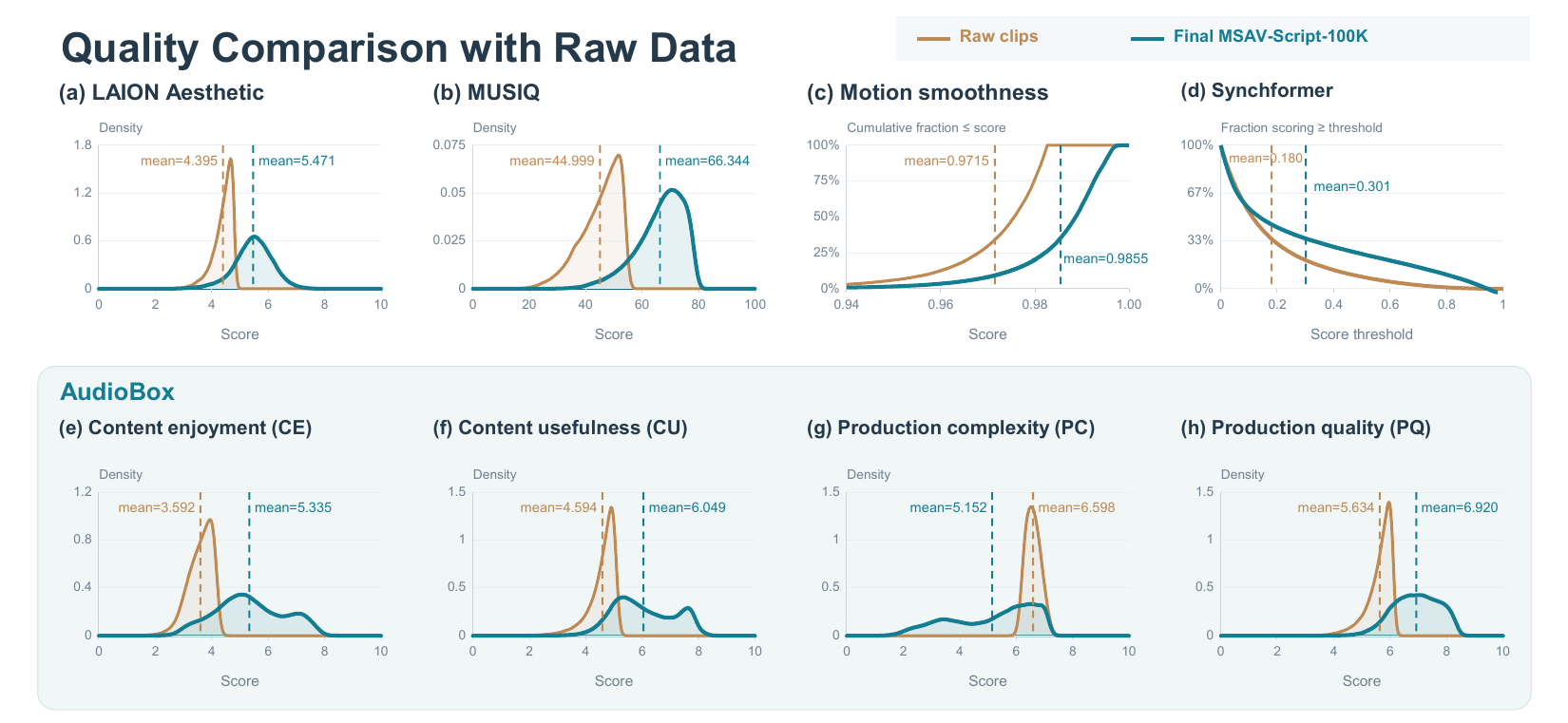}
    \vspace{-0.3 cm}
    \caption{\textbf{Quality assessment of MSAV-Script-100K
    against raw data.}
    The final dataset exhibits higher visual aesthetics,
    perceptual quality, and motion smoothness (a--c),
    stronger audio-visual synchronization (d), and improved
    audio enjoyment, usefulness, and production quality
    (e, f, h). Audio production complexity (g) characterizes
    differences in audio composition. 
    }
    \vspace{-0.2 cm}
    \label{fig:msav_script_100k_info}
\end{figure}
\subsection{Multi-Stage Quality Curation}

Our curation pipeline comprises perceptual and semantic screening
followed by tri-dimensional quality assessment, as shown in
Figure~\ref{fig:msav_script_pipeline}. The first stage removes clips with insufficient resolution,
undesirable visual or audio characteristics, or limited semantic
content; the second assesses visual, audio, and audio-visual quality jointly across complementary quality dimensions.

\paragraph{Perceptual and Semantic Screening.}
We first retain clips with a
resolution of at least 720p, ensuring  spatial quality
for subsequent assessment.

\textit{Visual quality filtering.} We evaluate six visual metrics:
sharpness, image entropy, flat-region proportion, texture statistics,
MUSIQ~\citep{ke2021musiq}, and RAFT~\citep{teed2020raft}-based
motion magnitude. We discard clips below P5 in sharpness, entropy,
or MUSIQ, above P95 in flat-region proportion or texture score,
or outside the P5--P95 motion range, removing degraded,
uninformative, near-static, and extreme-motion content.

\textit{Audio quality filtering.} We  screen audio using
silence ratio, clipping ratio, an amplitude-based signal-to-noise
proxy, and effective bandwidth, which characterize silence,
waveform saturation, amplitude contrast, and spectral coverage.

\textit{Semantic screening.} Since  adjacent shots are
not necessarily semantically coherent, we use
Qwen3.5-27B~\citep{qwen2026qwen35} to score each clip from
0 to 100 for semantic meaningfulness and retain clips scoring
at least 90. This visual-only assessment complements the preceding
low-level criteria with a content-level measure.

\paragraph{Tri-Dimensional Quality Assessment.}
Following screening, we assess the remaining clips along three
complementary axes: visual quality, audio quality, and audio-visual
synchronization, comprising five evaluation criteria for comprehensive
multi-modal quality assessment.


\textit{Visual quality.} We evaluate aesthetics with the
LAION Aesthetic Predictor~\citep{laion2022aesthetic} and motion
smoothness with AMT~\citep{li2023amt}, covering both frame-level
appearance and temporal quality across diverse visual contents and
motion patterns in the curated clips.

\textit{Audio quality.} We use AudioBox~\citep{tjandra2025audiobox}
to assess aesthetics and production quality, and
DNSMOS~\citep{reddy2021dnsmos} and NISQA~\citep{mittag2021nisqa}
to evaluate speech quality.

\textit{Audio-visual synchronization.} We use the Synchformer
zero-offset probability~\citep{iashin2024synchformerefficientsynchronizationsparse}
to measure temporal alignment between the visual and audio streams.
Finally, metric values are converted to within-source percentile
ranks and aggregated into five criterion-level subscores, whose
equal-weighted average defines the overall quality score.

\subsection{Dataset Statistics and Quality Analysis}

\paragraph{Dataset statistics.}
MSAV-Script-100K comprises 100,000 multi-shot audio-visual
clips, totaling approximately 237.4 hours.
Clips span approximately 6--12 seconds, with a mean duration
of 8.55 seconds and a median of 7.97 seconds.
Automatic shot counting identifies an average of 5.35 shots
per clip, with a median of five, providing compact sequences
with frequent visual transitions and continuous source audio.
The collection combines open-source videos with six categories
of YouTube content, covering diverse subjects and soundscapes.
Automatic audio tagging detects speech in 54.8\% of clips
and music in 49.7\%, with both tags present in 25.7\%.
These overlapping categories reflect the coexistence of
multiple auditory components within individual clips.
Detailed statistics and measurement protocols are provided
in Appendix~\ref{app:dataset_statistics}.

\paragraph{Quality analysis.}

Figure~\ref{fig:msav_script_100k_info} compares MSAV-Script-100K
with raw data in terms of visual quality, audio quality, and
audio-visual synchronization. For visual quality, the final dataset
achieves higher mean LAION aesthetic and MUSIQ scores, together with
better AMT-based motion smoothness than the raw-data reference set.
For audio quality, AudioBox reports higher content enjoyment,
content usefulness, and production quality, while production complexity
is reported separately as a descriptor of audio composition.
For audio-visual alignment, the mean Synchformer zero-offset probability
increases from 0.18 in the raw-data reference to 0.301 in
MSAV-Script-100K, indicating stronger temporal correspondence between
the visual and audio streams.




 \vspace{-0.1 cm}
 \vspace{-0.1 cm}
\section{Structured Stateful Relational Scripts}
\label{sec:ssrs}
 \vspace{-0.1 cm}
 \vspace{-0.1 cm}

\begin{figure}[t]
    \centering
    \includegraphics[width=\linewidth]{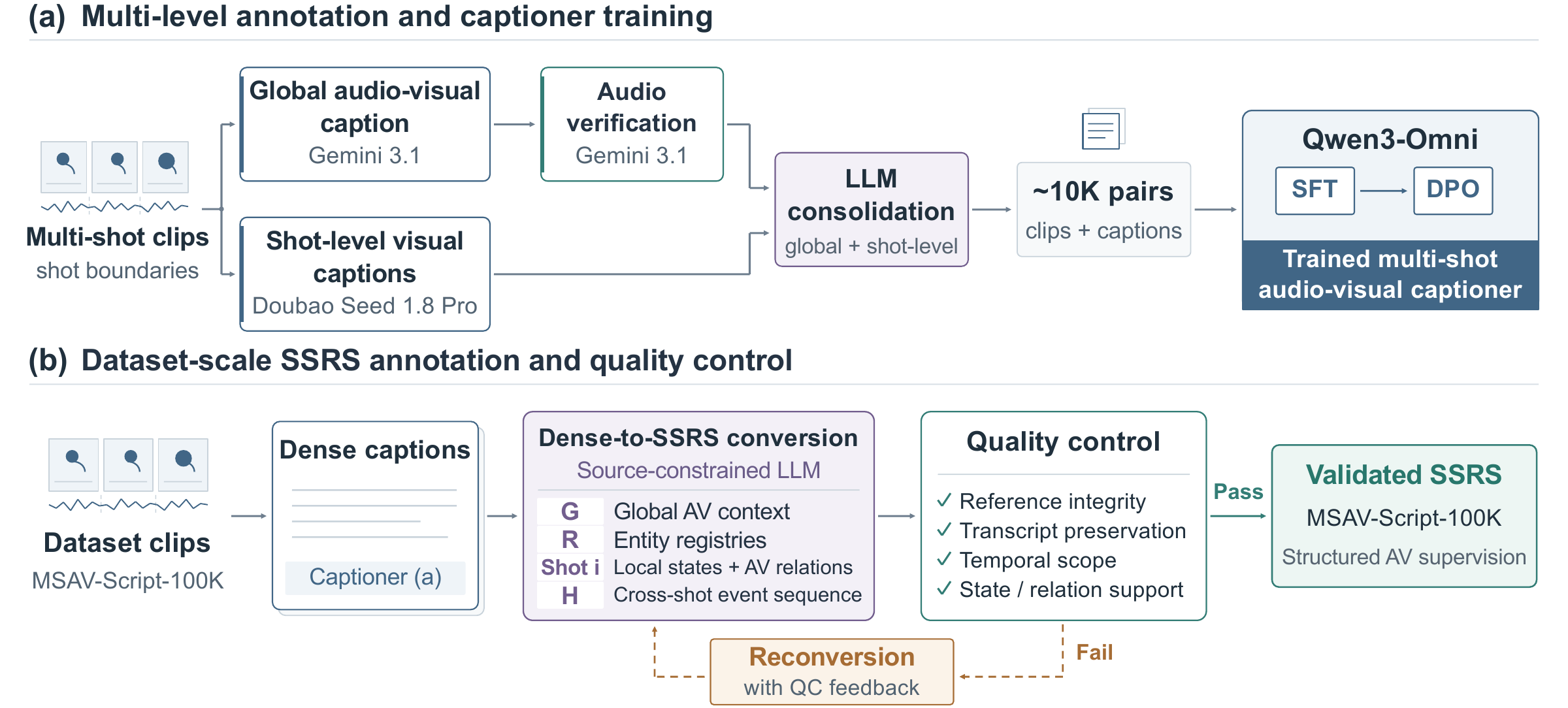}
    \vspace{-0.2 cm}
    \caption{\textbf{From content acquisition to structured supervision.}
\textbf{(a)} Multi-level expert annotations train a multi-shot
audio-visual captioner via SFT and DPO.
\textbf{(b)} Dense captions for MSAV-Script-100K undergo
source-constrained SSRS conversion and quality control.
}
\vspace{-0.2 cm}
    \label{fig:ssrs_pipeline}
\end{figure}
Our pipeline separates \emph{content acquisition} from
\emph{supervision organization}. We first develop a multi-shot
audio-visual captioner that produces comprehensive dense descriptions.
We then reorganize these descriptions into Structured Stateful
Relational Scripts (SSRS), exposing reusable identities,
shot-dependent states, and their audio-visual relations.
This separation makes caption structure a distinct design choice
rather than conflating it with a stronger description model.
Figure~\ref{fig:ssrs_pipeline} summarizes the two stages. Panel (a) illustrates the training pipeline for our multi-shot audio-visual captioner, while panel (b) presents the dense-to-SSRS conversion process.

\subsection{Multi-Shot Audio-Visual Captioner}
\label{sec:captioner}

\paragraph{Multi-level annotation.}
Capturing a multi-shot clip requires both a global account and
locally grounded details. Given a clip and its shot boundaries,
we use Gemini~3.1 to generate a global audio-visual caption and
Doubao Seed~1.8 Pro to describe each shot's visual content.
The former provides clip-level context, while the latter supplies
shot-specific subjects, actions, scene details, and camera presentation.
Gemini~3.1 subsequently re-examines the audio descriptions to check
auditory content. Finally, an LLM consolidates these outputs into
a comprehensive dense caption, integrating the available visual
and auditory information while retaining the shot sequence.

\paragraph{Captioner training.}
We annotate approximately 10K multi-shot clips with this pipeline
and use the resulting video--caption pairs to supervise
Qwen3-Omni~\citep{xu2025qwen3omni}.
After supervised fine-tuning, we apply direct preference
optimization~\citep{rafailov2023dpo} to obtain our captioner.
The trained model produces dense captions for the dataset;
the dense-to-SSRS conversion is performed afterward.
Thus, the expert annotation pipeline supplies captioner supervision,
whereas SSRS determines how the resulting content is organized
for generation.

\begin{figure}[t]
    \centering
    \includegraphics[width=\linewidth]{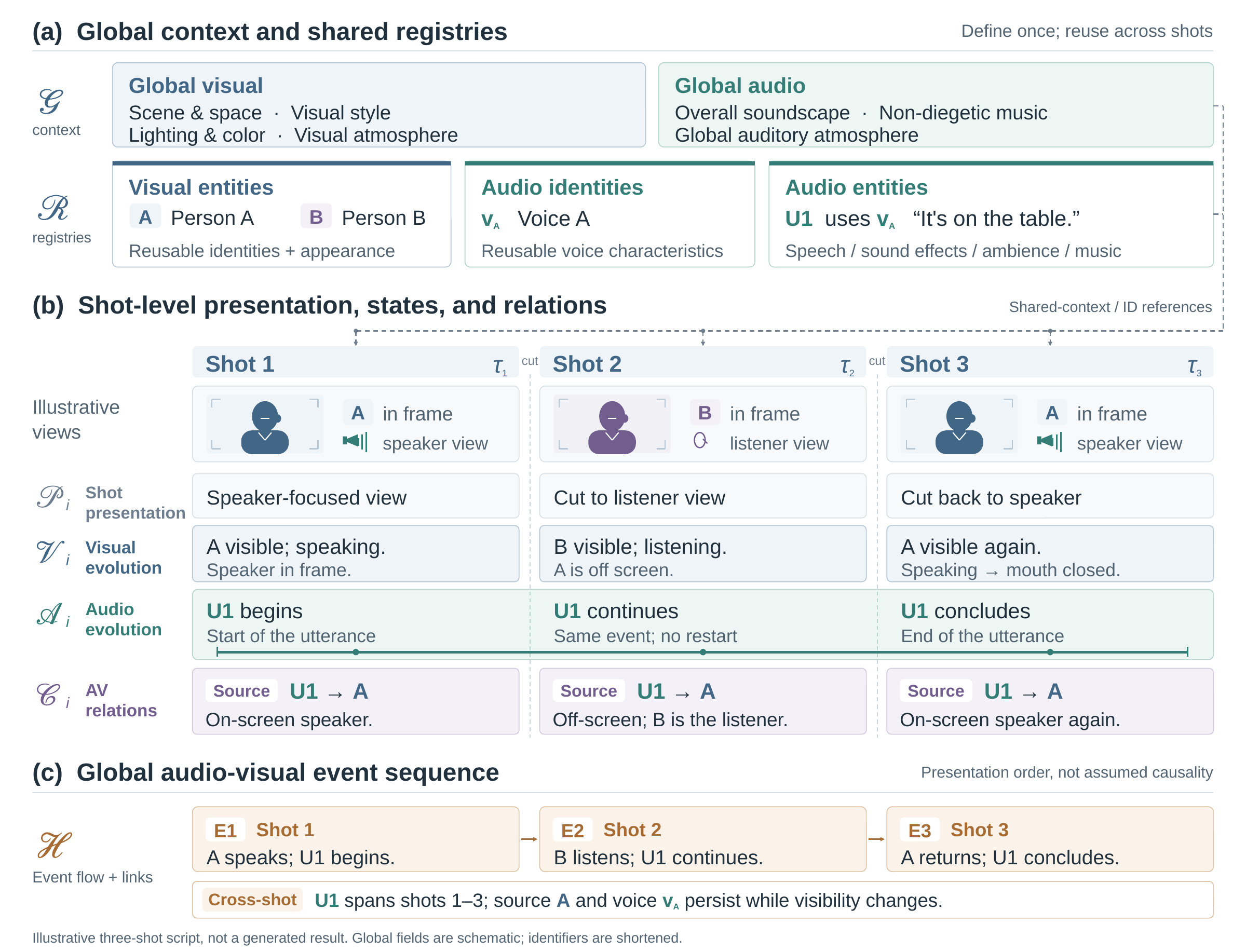}
    \vspace{-0.2 cm}
    \vspace{-0.1 cm}
    \caption{
    \textbf{SSRS distinguishes persistent identities from evolving states.}
Global context and shared registries \textbf{(a)} are separated
from shot-level presentation, states, and relations \textbf{(b)}.
The global event sequence \textbf{(c)} links events across cuts:
the same utterance continues despite changing speaker visibility,
highlighting that \emph{shot boundaries need not coincide with
event boundaries}.
    }
    \vspace{-0.2 cm}
    \vspace{-0.1 cm}
    \label{fig:ssrs_cross_shot}
\end{figure}

 \vspace{-0.1 cm}
\subsection{Representation Overview}
\label{sec:ssrs_overview}
 \vspace{-0.1 cm}
Dense prose often interleaves recurring identities, transient
attributes, actions, and sounds. SSRS assigns these factors
distinct semantic roles and temporal scopes. For a clip with
$K$ shots, we write
\begin{equation}
\mathcal{S}=
\left(\mathcal{G},\mathcal{R},
\{(\tau_i,\mathcal{P}_i,\mathcal{V}_i,
\mathcal{A}_i,\mathcal{C}_i)\}_{i=1}^{K},\mathcal{H}\right),
\label{eq:ssrs}
\end{equation}
where $\mathcal{G}$ is global context and $\mathcal{R}$ contains
shared registries. Each shot specifies its source time interval
$\tau_i$, cinematic presentation $\mathcal{P}_i$, visual states
and changes $\mathcal{V}_i$, auditory states and changes
$\mathcal{A}_i$, and cross-modal relations $\mathcal{C}_i$.
The global event sequence $\mathcal{H}$ organizes key events
and cross-shot connections. Figure~\ref{fig:ssrs_cross_shot} provides a concrete example of this representation.
The presentation fields retain transitions, shot size, view angle,
camera movement, local lighting, and frame details as specified
in the source caption.
This preserves cinematic instructions alongside the state--relation
factorization; scene layout and visible text are not discarded
merely because they do not participate in the main event sequence.

\subsection{Global Context and Entity Registries}
\label{sec:ssrs_registries}

\paragraph{Global context.}
Global visual context comprises four complementary aspects:
\emph{scene and space} describe the shared environment and
spatial layout; \emph{visual style} characterizes the overall
aesthetic; \emph{lighting and color} specify shared illumination
and the color palette; and \emph{visual atmosphere} captures
the overall mood of the scene.
Global audio comprises three aspects:
\emph{overall soundscape} describes environmental sounds,
action sounds, and non-verbal vocalizations, including their
spatial characteristics and layering;
\emph{non diegetic music} summarizes the acoustic properties
of audience-only scoring; and
\emph{global auditory atmosphere} captures the overall
auditory mood explicitly. 

\paragraph{Shared registries.}
SSRS organizes reusable content into three registries.
The \emph{visual entity registry} defines entity identities
and appearance attributes; the \emph{audio identity registry}
records voice characteristics; and the \emph{audio entity registry}
specifies speech, sound effects, ambience, and music, including
 content and acoustic properties.
Transcribed speech entries preserve the verbatim transcript.
Speech entries reference a reusable voice identity when
supported by the source, and untranscribed speech is
explicitly represented without inventing a transcript.
Unique identifiers allow shot descriptions and the global
event sequence to reference these entries without repeating
their definitions.


\subsection{Shot-Level Presentation, States, and Relations}
\label{sec:ssrs_local}
For each shot with source time interval $\tau_i$, SSRS separately
represents cinematic presentation, visual evolution, auditory
evolution, and cross-modal relations.
$\mathcal{P}_i$ describes transitions, shot size, view angle,
camera movement, local lighting and visual style, and frame
details such as spatial layout and visible text.
$\mathcal{V}_i$ records entity positions, poses, expressions,
actions, interactions, and supported changes. Actions with
unspecified final states are retained without adding unobserved
transitions.
$\mathcal{A}_i$ references audio entities and describes their
occurrence, continuation, repetition, overlap, and local changes,
while transcripts and stable acoustic descriptions remain
in the registries.
$\mathcal{C}_i$ links audio identifiers to visual entities or
concise action anchors, specifying source attribution,
synchronization, temporal order, duration, overlap, and causation
when supported by the source. These relations are distinct:
co-occurrence does not imply synchronization, nor does
synchronization imply causality. For example, chewing and its
associated sound are represented as a visual action and an audio
entity, with their synchronization expressed separately
in $\mathcal{C}_i$.

\subsection{Cross-Shot Event Organization}
\label{sec:ssrs_cross_shot}

The global audio-visual event sequence $\mathcal{H}$ organizes
key actions, state changes, and auditory events into a main
event flow $E_1 \rightarrow E_2 \rightarrow \cdots$.
Each node links to local descriptions through existing
identifiers and concise action references, with its corresponding
shot or shot range. The ordering follows video presentation
without assuming causality or chronological story time.
Additional source-supported relations specify simultaneity,
persistence, cross-shot dependencies, and causal triggers
without repeating local details.
This organization captures both visual state evolution and
auditory continuity across cuts. For example, pouring into the
same jar may continue across shots before the jar is repositioned.
Likewise, an utterance may continue after a cut from the speaker
to a listener, retaining the
same speech event and voice identity while the speaker's
visibility changes. Thus, shot boundaries need not coincide
with event boundaries, and $\mathcal{H}$ explicitly organizes
both continuity and intended change across shots.

\subsection{Dense-to-Script Conversion and Quality Control}
\label{sec:ssrs_conversion}

We formulate dense-to-SSRS conversion as source-constrained
reorganization. An LLM maps source-caption information to
global context, shared registries, shot-level fields, and the
global event sequence according to semantic roles and temporal
scopes. The conversion is designed to preserve source facts,
shot boundaries, source-provided timing, verbatim dialogue and
visible text, as well as uncertainty and explicit negation.
To reduce redundancy, each fact has a primary location,
with identifier-based references avoiding repeated definitions.
The conversion specification requires checks for reference
integrity, transcript preservation, temporal scope, and
unsupported state or relation assertions. These checks help preserve source fidelity and structural consistency, supporting more reliable supervision for multi-shot audio-visual generation.



\section{Experiments}
\label{sec:experiments}

We organize the evaluation around two complementary questions: whether SSRS improves general multi-shot audio-visual generation, and whether it improves the specific state and relational controls that motivate its design. Public MSAVBench~\citep{msavbench} comparisons address the former; controlled evaluation and representation ablations address the latter. 
\vspace{-0.1 cm}
\subsection{Experimental Setup}
\vspace{-0.1 cm}
\label{sec:exp_setup}
\paragraph{Backbone and training controls.}
Our controlled comparisons use LTX-2.3 with three primary configurations: the pretrained backbone, dense-caption supervised fine-tuning, and SSRS-supervised fine-tuning. Dense and SSRS supervision are derived from the same training clips and source descriptions. To isolate the representation, the two fine-tuned configurations use identical data splits, initialization, trainable parameters, optimization schedules, and numbers of updates.  

\paragraph{Evaluation and metrics.}
We evaluate text-to-audio-video (T2AV) generation on a fixed,
text-only subset of MSAVBench~\citep{msavbench} comprising
150 short-video prompts. All methods are scored against the
same canonical descriptions. We retain 14 metrics that do not
require reference media, covering three levels:
\emph{global}---narrative coherence (Narr), lip synchronization
(Lip), sound attribution (Attr), audio-visual synchronization
(Sync), and visual question-answering accuracy (VQ);
\emph{cross-shot}---layout consistency (C-Layout), visual
consistency (VC), music consistency (Mus), and speaker-timbre
consistency (Spk); and
\emph{intra-shot}---layout--text alignment (I-Layout), camera
adherence (Cam), perceptual audio quality (PQ), text rendering
(OCR), and spoken-content fidelity (WER-S).
We use Gemini~3.5~Flash for MLLM-based judgments,
together with metric-specific expert models and tools.
All reported scores lie in $[0,1]$, with higher values
indicating better performance, using metric-specific
scoring and normalization rules.


\begin{table}[t]
\centering
\caption{Comparison on the 150-prompt short-video T2AV subset of
MSAVBench. All scores are normalized to $[0,1]$, where higher
values indicate better performance. Both Dense SFT and SSRS SFT
are fine-tuned from the pretrained LTX-2.3 model.
C-L and I-L denote C-Layout and I-Layout, respectively, while
WER-S is computed as $\mathrm{WER\text{-}S}=\operatorname{clip}(1-e,0,1)$,
where $e$ denotes the applicable transcription error rate
(WER or CER).
Best and second-best results are highlighted in
\textbf{bold} and \underline{\textit{underlined italics}},
respectively.  }
\label{tab:main_comparison}

\small
\setlength{\tabcolsep}{2.0pt}
\renewcommand{\arraystretch}{1.0}

\resizebox{1.0\linewidth}{!}{%
\begin{tabular}{l *{5}{c} *{4}{c} *{5}{c}}
\toprule

& \multicolumn{5}{c}{\textbf{Global}}
& \multicolumn{4}{c}{\textbf{Cross-shot}}
& \multicolumn{5}{c}{\textbf{Intra-shot}} \\

\cmidrule(lr){2-6}
\cmidrule(lr){7-10}
\cmidrule(lr){11-15}


\textbf{Method}
& Narr$\uparrow$
& Lip$\uparrow$
& Attr$\uparrow$
& Sync$\uparrow$
& VQ$\uparrow$
& C-L$\uparrow$
& VC$\uparrow$
& Mus$\uparrow$
& Spk$\uparrow$
& I-L$\uparrow$
& Cam$\uparrow$
& PQ$\uparrow$
& OCR$\uparrow$
& WER-S$\uparrow$ \\
\midrule
Ovi
& 0.3288 & 0.0605 & 0.2864 & 0.8290 & 0.5434
& 0.4333 & 0.6824 & 0.4978 & 0.3094
& 0.9136 & 0.5316 & 0.5733 & \underline{\textit{0.2222}} & 0.0065 \\

JavisDiT++
& 0.4932 & 0.0084 & 0.0668 & 0.8073 & 0.6499
& 0.6731 & 0.6747 & 0.5055 & \textbf{0.4651}
& 0.6288 & 0.6245 & 0.4981 & 0.1379 & 0.0016 \\

JavisGPT
& 0.2170 & 0.0022 & 0.0101 & 0.7780 & 0.3813
& 0.6400 & 0.7038 & 0.4992 & 0.0659
& 0.8333 & 0.4281 & 0.5575 & 0.0053 & 0.0028 \\

ShotStream+HF
& 0.6844 & 0.1384 & 0.2762 & 0.7863 & 0.5943
& \underline{\textit{0.8704}} & 0.8720 & 0.4722 & 0.1860
& 0.9151 & 0.6377 & \textbf{0.6641} & 0.1734 & 0.0033 \\

LTX-2.5
& 0.7883 & 0.1242 & 0.2500 & 0.7727 & 0.7269
& 0.8152 & 0.8664 & 0.4673 & 0.1418
& 0.8902 & 0.7529 & \underline{\textit{0.6416}} & \textbf{0.2324} & 0.7226 \\

\midrule

Base-LTX-2.3
& 0.8076 & 0.1138 & 0.2061 & 0.7840 & 0.8033
& 0.7301 & \textbf{0.8905} & 0.4666 & 0.1811
& \underline{\textit{0.9167}} & 0.7648 & 0.6235 & 0.1829 & 0.7262 \\

\midrule

Dense SFT
& \underline{\textit{0.8222}}
& \underline{\textit{0.2284}}
& \underline{\textit{0.5104}}
& \underline{\textit{0.8450}}
& \underline{\textit{0.8202}}
& 0.8409
& \underline{\textit{0.8756}}
& \underline{\textit{0.7217}}
& 0.4002
& \textbf{0.9722}
& \underline{\textit{0.8027}}
& 0.6244
& 0.1389
& \underline{\textit{0.7319}} \\

\textbf{SSRS SFT}
& \textbf{0.8467}
& \textbf{0.2468}
& \textbf{0.5265}
& \textbf{0.8573}
& \textbf{0.8376}
& \textbf{0.9167}
& 0.8631
& \textbf{0.7523}
& \underline{\textit{0.4006}}
& 0.8750
& \textbf{0.8321}
& 0.6011
& 0.2118
& \textbf{0.7543} \\

\bottomrule
\end{tabular}%
}
\vspace{-0.4 cm}
\end{table}

\vspace{-0.2 cm}

\subsection{Comparison on MSAVBench}
\label{sec:exp_msav}
\vspace{-0.2 cm}
We compare against representative open-source audio-video and
multi-shot generation systems, including Ovi~\citep{ovi},
JavisDiT++~\citep{liu2026javisditpp},
JavisGPT~\citep{liu2025javisgpt},
ShotStream+HF~\citep{luo2026shotstream,shan2025hunyuanvideofoley},
and LTX-2.5~\citep{ltx25}. Since Ovi, JavisDiT++, and JavisGPT
do not natively support multi-shot generation, we follow the
MSAVBench protocol~\citep{msavbench} and generate each shot
independently before concatenating them into a multi-shot sequence.
We also report the pretrained LTX-2.3 backbone~\citep{ltx2} to
disentangle the effects of data adaptation and structured supervision. As shown in Table~\ref{tab:main_comparison}, dense-caption
fine-tuning on MSAV-Script-100K improves most metrics over the
pretrained backbone, demonstrating the effectiveness of our
multi-shot audio-visual training data. Under matched training
conditions, SSRS further improves 11 of the 14 metrics over Dense
SFT, with consistent gains across all global metrics and notable
improvements in cross-shot and instruction-following measures.
This indicates that explicitly organizing identities, evolving
states, and audio-visual relations provides benefits beyond data
adaptation alone.
Compared with external systems, SSRS SFT achieves the best
performance on 9 of the 14 metrics. Some metrics, including VC,
I-Layout, and PQ, remain below Dense SFT, suggesting that stronger
structured control does not uniformly improve every aspect of
visual consistency and perceptual quality.










\begin{table}[t]
\centering
\caption{\textbf{Controlled evaluation of state and relational following.}
$T$ denotes target satisfaction and $P$ denotes preservation rate.
All values are percentages. $\Delta$ denotes SSRS SFT minus Dense SFT.}
\label{tab:controlled_states_relations}

\small
\setlength{\tabcolsep}{4.5pt}
\renewcommand{\arraystretch}{1.05}

\resizebox{0.85 \linewidth}{!}{%
\begin{tabular}{l  cc cc cc}
\toprule
& \multicolumn{2}{c}{Dense SFT}
& \multicolumn{2}{c}{SSRS SFT}
& \multicolumn{2}{c}{$\Delta$} \\
\cmidrule(lr){2-3}
\cmidrule(lr){4-5}
\cmidrule(lr){6-7}

\textbf{Category} 
& $T\uparrow$ & $P\uparrow$
& $T\uparrow$ & $P\uparrow$
& $\Delta T$ & $\Delta P$ \\
\midrule

Visual-state evolution (VS)
& 20.00 & \textbf{62.50}
& \textbf{60.00} & 42.50
& +40.00 & -20.00 \\

Cross-shot audio continuity (AC)
& 0.00 & 59.38
& \textbf{50.00} & \textbf{62.50}
& +50.00 & +3.12 \\

Non-speech source binding (NS)
& 5.00 & 45.00
& \textbf{45.00} & \textbf{60.00}
& +40.00 & +15.00 \\

Cross-shot event order (EO)
& 0.00 & 50.00
& \textbf{40.00} & \textbf{55.00}
& +40.00 & +5.00 \\

On-/off-screen speaker binding (SB)
& 7.14 & 25.00
& \textbf{42.86} & \textbf{42.86}
& +35.71 & +17.86 \\

\midrule
\textbf{Overall}
& 7.50 & 49.38
& \textbf{48.75} & \textbf{52.50}
& \textbf{+41.25} & \textbf{+3.12} \\

\bottomrule
\end{tabular}%
}
\vspace{-0.3 cm}
\end{table}

\subsection{Controlled Evaluation of States and Relations}
\label{sec:exp_controlled}

MSAVBench measures general generation quality but does not directly
evaluate the state transitions and event relations targeted by SSRS.
We therefore construct a controlled diagnostic set covering five
aspects: visual-state evolution (VS), cross-shot audio continuity
(AC), non-speech sound-source binding (NS), cross-shot event order
(EO), and on-/off-screen speaker binding (SB). We evaluate 40
matched prompt instances for Dense SFT and SSRS SFT. Each output is
assessed by Gemini~3.5~Flash with native audio using two target
questions and four preservation questions, yielding target
satisfaction ($T$) and preservation ($P$) rates. We further perform manual verification of the model-based judgments to ensure the accuracy and reliability of the evaluation results.
As shown in Table~\ref{tab:controlled_states_relations}, SSRS
substantially increases overall $T$ from 7.50\% to 48.75\%
(+41.25 pp), while maintaining comparable preservation
(49.38\% vs.\ 52.50\%). Target satisfaction improves across all
five categories, with gains of 35.71--50.00 pp, while $P$ improves
in four categories. The exception is visual-state evolution, where
higher target adherence is accompanied by lower preservation.
Overall, these results indicate that \textbf{SSRS substantially improves
adherence to intended state and relational changes without
degrading preservation of shared constraints overall}.

\begin{table}[t]
\centering
\caption{Representation ablations under matched training and
evaluation settings. All scores are normalized to $[0,1]$,
where higher values indicate better performance.}
\label{tab:representation_ablation}
\resizebox{\linewidth}{!}{%
\begin{tabular}{l ccc ccc cc}
\toprule
& \multicolumn{3}{c}{\textbf{ Cross-shot Organization}}
& \multicolumn{3}{c}{\textbf{Audio-Visual Relations}}
& \multicolumn{2}{c}{\textbf{Control Metrics}} \\
\cmidrule(lr){2-4}
\cmidrule(lr){5-7}
\cmidrule(lr){8-9}

\textbf{Representation}
& Narr$\uparrow$
& Mus$\uparrow$
& Spk$\uparrow$
& Lip$\uparrow$
& Attr$\uparrow$
& Sync$\uparrow$
& VQ$\uparrow$
& Cam$\uparrow$ \\
\midrule

Full SSRS
& \underline{\textit{0.8467}} & \textbf{0.7523} & \underline{\textit{0.4006}}
& \underline{\textit{0.2468}} & \textbf{0.5265} & \textbf{0.8573}
& \textbf{0.8376} & \underline{\textit{0.8321}} \\
w/o Audio-Visual Relations

& \textbf{0.8622} & 0.7131 & \textbf{0.4268}
& 0.2374 & \underline{\textit{0.5110}} & 0.8387
& 0.8137 & \textbf{0.8420} \\

w/o Global  Registries and Event
& 0.8289 & \underline{\textit{0.7168}} & 0.3801
& \textbf{0.2530} & 0.5003 & \underline{\textit{0.8477}}
& \underline{\textit{0.8189}} & 0.8218 \\
\bottomrule
\end{tabular}%
}
\vspace{-0.4 cm}
\end{table}

\subsection{Representation Ablations}
\label{sec:exp_ablation}
\vspace{-0.1 cm}
We ablate two key components of SSRS under matched training and
evaluation settings: the explicit audio-visual relation fields and
the global entity registries and event. As shown in
Table~\ref{tab:representation_ablation}, removing audio-visual
relations degrades sound attribution, synchronization, music
consistency, and visual-question accuracy, although a few individual
metrics slightly improve. 
Removing global registries and event organization
further weakens cross-shot organization and identity-related
consistency, indicating the importance of reusable entity references
across shots. Overall, the full SSRS provides the most balanced
performance across cross-shot organization, audio-visual relations,
and control metrics, supporting the complementary roles of explicit
relational modeling and global  organization.

\vspace{-0.2 cm}
\section{Conclusion}
\label{sec:conclusion}
\vspace{-0.2 cm}
We presented MSAV-Script-100K and Structured Stateful Relational
Scripts (SSRS) for multi-shot audio-visual generation. SSRS
organizes persistent identities, evolving visual and auditory states,
cross-modal relations, and cross-shot event continuity into explicit
generation supervision. Experiments on LTX-2.3 show that our dataset
improves general multi-shot audio-visual generation, while SSRS further
enhances state and relational adherence under matched training
conditions. A remaining limitation is that SSRS inherits the coverage
and accuracy of its source captions, motivating stronger audio-visual
captioning and verification in future work.

\subsection*{AI use statement}

We used ChatGPT and OpenAI Codex to assist with English polishing,
Chinese--English translation, manuscript organization, experimental
design discussion, \LaTeX{} editing, figure design, and implementation
support. All AI-assisted text, experimental plans, code, and figures
were manually reviewed, revised, and verified by the authors before
being incorporated into the paper. Generative AI systems that constitute
part of our research methodology, including those used for annotation,
captioning, and evaluation, are explicitly described in the corresponding
sections of the paper. The authors take full responsibility for the final
content, claims, experiments, code, and artifacts presented in this work.

\subsection*{Ethics statement}
MSAV-Script-100K is constructed from publicly accessible video
sources, including existing open datasets and publicly available
YouTube videos collected for multi-shot audio-visual research.
The dataset is intended solely for academic and non-commercial
research. We respect the copyright, privacy, and personality rights
associated with the source content, and do not intend the dataset
for biometric identification, surveillance, or other privacy-invasive
applications. Where redistribution of original media is restricted,
we will release annotations, metadata, timestamps, and source references
rather than redistribute the underlying content without permission.
We will also provide a mechanism through which content owners or
featured individuals may request removal of relevant samples from
future releases. All released materials will be accompanied by clear
documentation of their intended research use and applicable licensing
conditions. Our annotations, metadata, and derived textual resources will be
released under the CC BY-NC-SA 4.0 license, while the redistribution
of source videos will follow their original licenses and platform
terms.
\subsection*{Reproducibility statement}

We provide detailed descriptions of the dataset construction,
multi-shot audio-visual captioning pipeline, SSRS representation,
training protocol, evaluation metrics, and controlled diagnostic
experiments throughout the main paper and appendix. Implementation
details, prompts, data-processing procedures, training configurations,
and evaluation protocols are documented to facilitate reproduction.
Upon publication, we plan to release the MSAV-Script-100K annotations,
SSRS data, training and evaluation code, model checkpoints, and the
necessary configuration files, subject to the licensing constraints
of the original video sources. These materials are intended to support end-to-end reproduction of
our data preparation, SSRS conversion, model training, and evaluation.

\bibliography{iclr2027_conference}
\bibliographystyle{iclr2027_conference}


\clearpage
\appendix

\section*{Appendix Contents}

\begingroup
\small
\setlength{\parindent}{0pt}

\newcommand{\appcontentsentry}[2]{%
  \noindent
  \hyperref[#1]{%
    \makebox[1.8em][l]{\ref*{#1}}#2%
  }%
  \nobreak\dotfill\nobreak
  \hyperref[#1]{\pageref*{#1}}\par
  \vspace{0.7em}%
}

\appcontentsentry{app:dataset}
  {Data Sources and Multi-Shot Construction}

\appcontentsentry{app:dataset_statistics}
  {Dataset Statistics}

\appcontentsentry{app:dataset_filtering}
  {Filtering and Final Selection Details}

\appcontentsentry{app:qualitative}
  {Qualitative Comparison and Analysis}

\appcontentsentry{app:controlled_eval}
  {Details of the Controlled State and Relation Evaluation}

\appcontentsentry{sec:Training}
  {Training Details}

\appcontentsentry{app}
  {Preference Optimization of the Captioner}

\appcontentsentry{sec:appendix_ssrs_conversion}
  {SSRS Conversion and Quality Control}

\appcontentsentry{sec:appendix_evaluation}
  {Evaluation Metrics and Implementation Details}

\endgroup

\clearpage

\section{Data Sources and Multi-Shot Construction}
\label{app:dataset}

\begin{figure}[h]
    \centering
    \includegraphics[width=\linewidth]{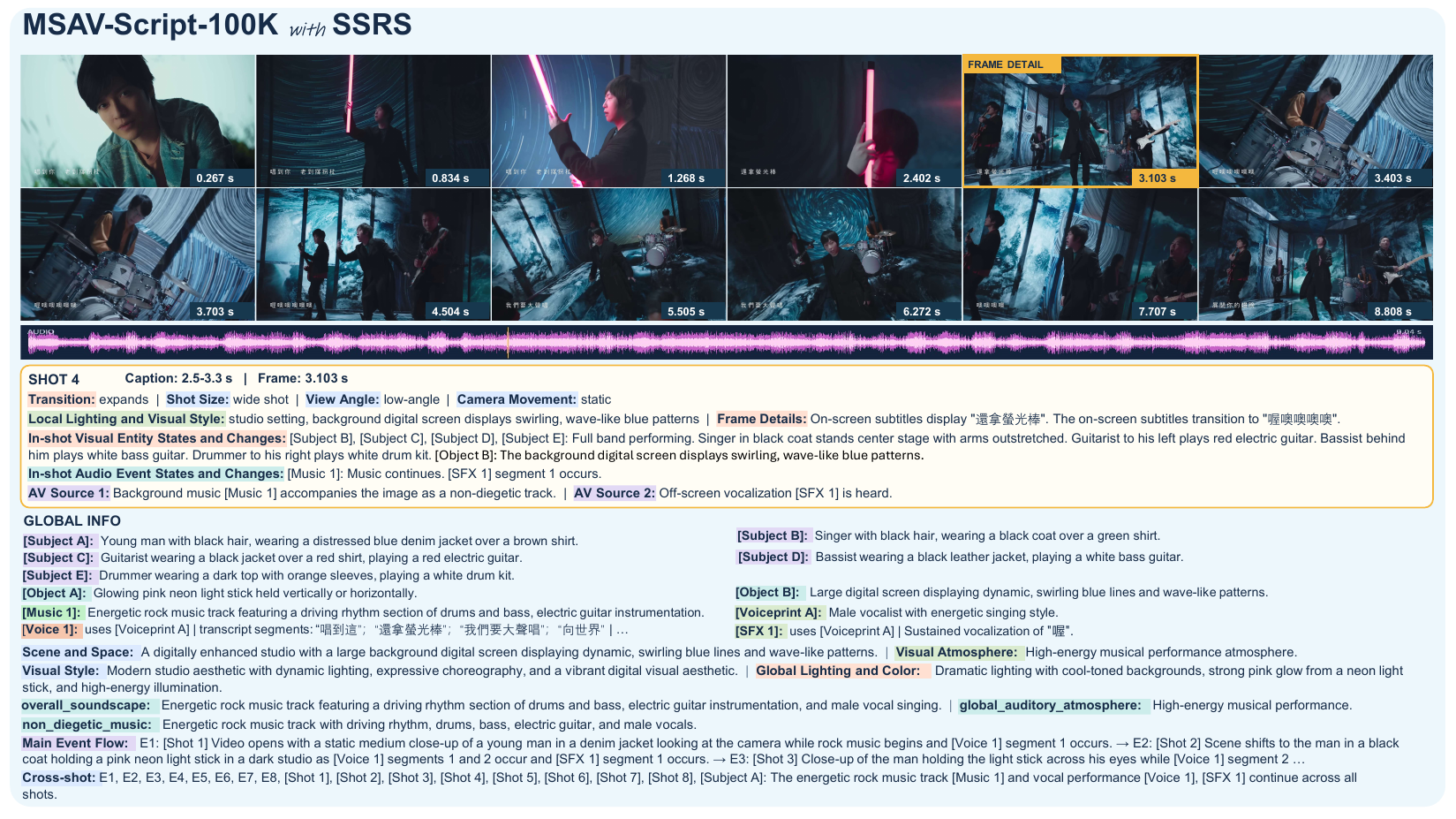}
    \caption{\textbf{An example from MSAV-Script-100K with SSRS annotations.}
    Sampled video frames and the audio waveform are shown together with shared entity and voice registries, global audio-visual context, shot-local states and source relations, and a cross-shot event sequence. The highlighted panel shows a shot-level annotation excerpt.}
    \label{fig:dataset_demo}
\end{figure}

This appendix details the source collection and temporal construction of MSAV-Script-100K. Appendix~\ref{app:dataset_statistics} reports dataset composition and temporal, visual, and audio statistics, and Appendix~\ref{app:dataset_filtering} describes screening and final selection. The processing unit is a contiguous audio-visual clip containing multiple shots, paired with a dense description and its SSRS representation.

\subsection{Data Sources and Content Coverage}
\label{app:dataset_sources}

Our source pool combines FineVideo, MiraData, VCapAV, and VidMuse (VidMuse-V2M-Dataset) with videos collected from YouTube. The YouTube collection covers six categories: food, music videos, animations, documentaries, sports events, and esports events. These categories broaden the range of visual activities and auditory content beyond cinematic dialogue. For example, food preparation provides action-associated sounds, musical performances combine recurring subjects with sustained music, and sports and esports footage includes commentary over changing views. These examples describe the intended coverage of the collection; individual clips need not contain every type of interaction associated with their category.

The collection targets short sequences with naturally occurring shot transitions and their corresponding audio. Native audio preserves the relationship between the original visual editing and the sound track, including cases in which narration, music, ambience, or an utterance continues across a cut. Accordingly, a change of shot does not require a change of speaker, sound source, or auditory event.

\subsection{Temporal Construction of Multi-Shot Clips}
\label{app:dataset_segmentation}

We use TransNet V2~\citep{soucek2020transnetv2effectivedeep} to detect shot boundaries in source videos. FFmpeg cuts the corresponding audiovisual segments at the resulting clip boundaries. Consecutive shots are grouped into clips of approximately 6--12 seconds. Each clip preserves the original shot order and the audio associated with the same continuous source interval, allowing speech, music, and ambience to continue across visual cuts.

Let $b_0<b_1<\cdots<b_K$ denote the boundaries of $K\geq2$ consecutive shots in the source timeline. A clip spanning $[b_0,b_K)$ has duration $b_K-b_0$, with clip-local shot intervals
\begin{equation}
    \tau_i=[b_{i-1}-b_0,\,b_i-b_0),\qquad i=1,\ldots,K.
    \label{eq:app_shot_intervals}
\end{equation}
These timestamps express visual segmentation relative to the clip start. An auditory event may span several intervals $\tau_i$; its boundaries are not required to coincide with visual cuts.

\section{Dataset Statistics}
\label{app:dataset_statistics}

\subsection{Dataset Composition}
\label{app:dataset_composition}

All statistics in this section refer to the final MSAV-Script-100K. Table~\ref{tab:app_dataset_overview} summarizes its scale and temporal structure; Tables~\ref{tab:app_source_composition} and~\ref{tab:app_resolution_distribution} report source composition and short-side resolution distribution. Counts and percentages use clips as the counting unit.

\begingroup
\setlength{\intextsep}{8pt}
\setlength{\abovecaptionskip}{5pt}
\setlength{\belowcaptionskip}{2pt}
\begin{figure}[h]
    \centering
    \begin{minipage}[c]{0.49\linewidth}
        \centering
        \includegraphics[width=\linewidth]{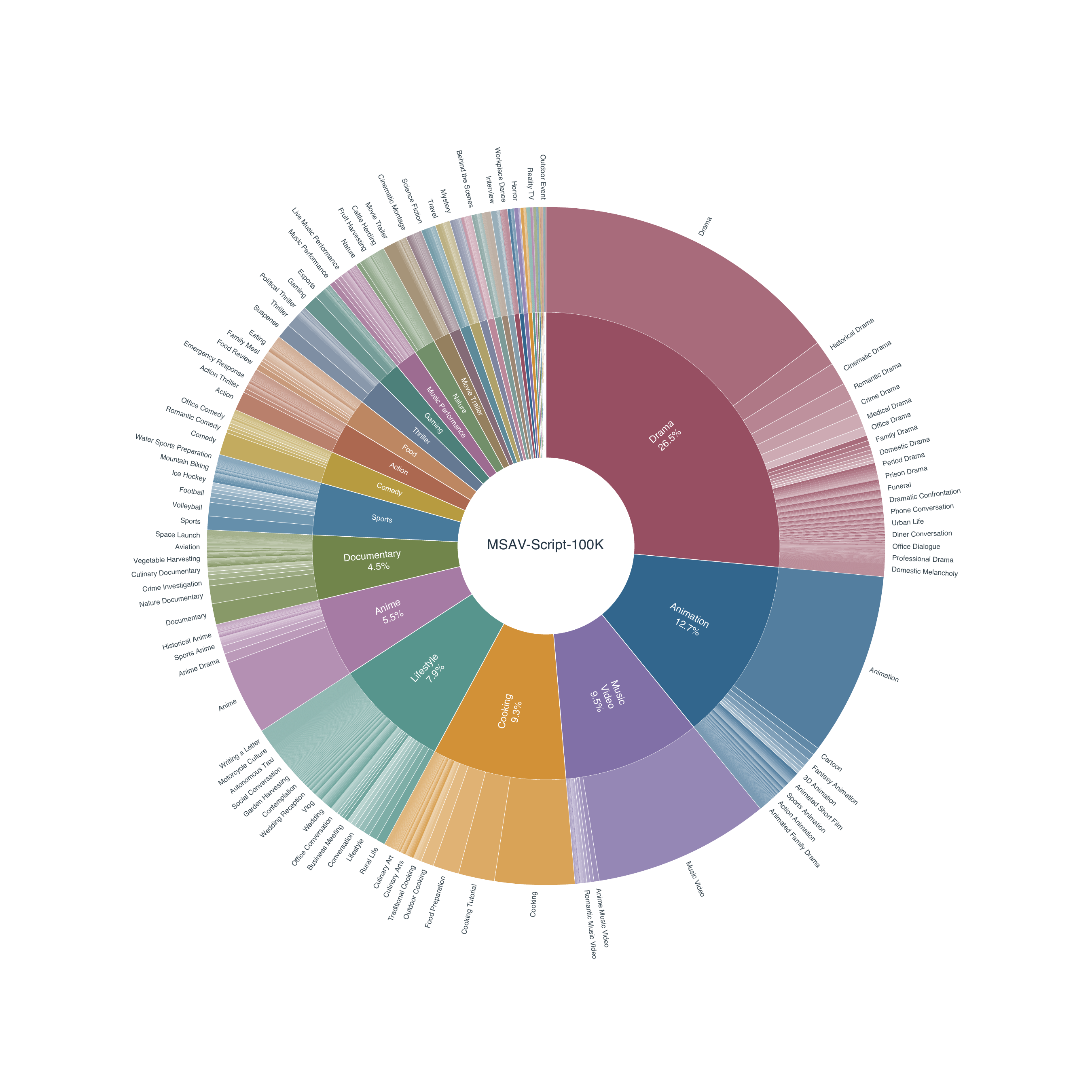}
    \end{minipage}
    \hfill
    \begin{minipage}[c]{0.49\linewidth}
        \centering
        \includegraphics[
            width=\linewidth
        ]{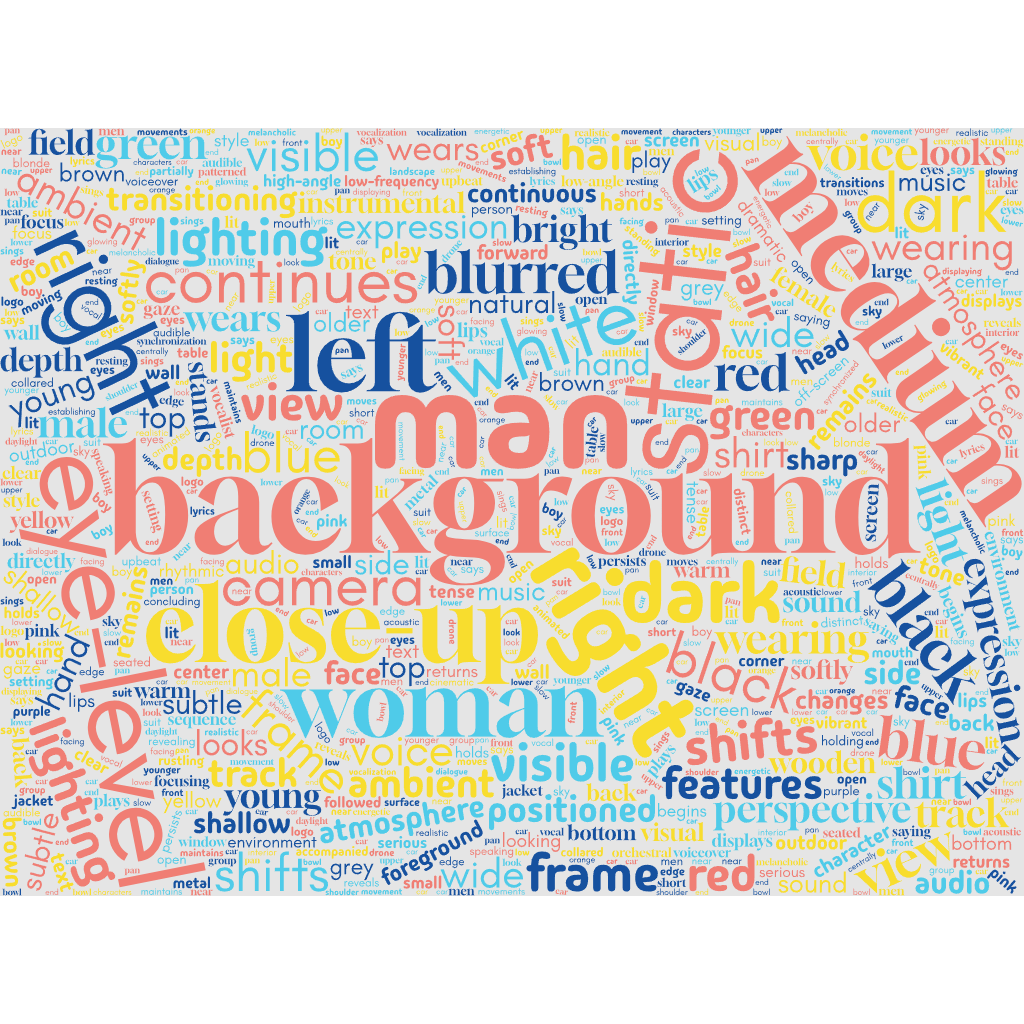}
    \end{minipage}
    \caption{Thematic taxonomy (left) of MSAV-Script-100K and word cloud (right) of SSRS, illustrating the dataset's topic
    hierarchy and frequently occurring words in its captions.}
    \label{fig:app_dataset_visualizations}
\end{figure}

\begin{table}[h]
\centering
\small
\caption{Dataset statistics for the selected 100,000 clips. Audio tags and shot counts are automatic estimates. Audio categories may overlap.}
\label{tab:app_dataset_overview}
\begin{tabularx}{\linewidth}{@{}lXl@{}}
\toprule
Group & Statistic & Value \\
\midrule
Scale & Number of selected clips & 100,000 \\
      & Total duration in hours & 237.426 \\
Sources & Clip counts by source & Table~\ref{tab:app_source_composition} \\
Temporal structure & Mean clip duration (s) & 8.547 \\
                   & Median clip duration (s) & 7.966 \\
                   & Clip duration range (s) & 6.083--12.040 \\
                   & Detected shots per clip: mean, median, and range & 5.346 / 5 / 2--75 \\
Visual properties & Short-side resolution & Table~\ref{tab:app_resolution_distribution} \\
Audio content & Automatically tagged clips by sound category & \shortstack[l]{Speech: 54.781\%\\Music: 49.737\%\\Other: 4.821\%} \\
\bottomrule
\end{tabularx}
\end{table}

\paragraph{Duration measurements.}
The selected clips contain approximately 237.426 hours of video. Their mean and median durations are 8.547 and 7.966 seconds, respectively, with a range of 6.083--12.040 seconds.


\begin{table}[!htp]
\centering
\begin{minipage}[t]{0.48\linewidth}
\vspace{0pt}
\setlength{\tabcolsep}{3pt}
\centering
\small
\caption{Source composition of the final 100,000 clips.}
\label{tab:app_source_composition}
\begin{tabularx}{\linewidth}{@{}Xr@{}}
\toprule
Source & Clips \\
\midrule
YouTube collections & 74,114 \\
MiraData & 23,950 \\
VidMuse & 918 \\
FineVideo & 518 \\
VCapAV & 500 \\
\midrule
Total & 100,000 \\
\bottomrule
\end{tabularx}
\end{minipage}\hfill
\begin{minipage}[t]{0.48\linewidth}
\vspace{0pt}
\setlength{\tabcolsep}{3pt}
\centering
\small
\caption{Resolution distribution of MSAV-Script-100K.}
\label{tab:app_resolution_distribution}
\begin{tabularx}{\linewidth}{@{}Xrr@{}}
\toprule
Short side (pixels) & Clips & Share (\%) \\
\midrule
720--1,079 & 51,832 & 51.832 \\
1,080--1,439 & 47,907 & 47.907 \\
1,440--2,159 & 147 & 0.147 \\
$\geq 2{,}160$ & 114 & 0.114 \\
\midrule
Total & 100,000 & 100.000 \\
\bottomrule
\end{tabularx}
\end{minipage}
\end{table}

\paragraph{Automatic audio tagging.}

We apply an AudioSet-trained Audio Spectrogram Transformer (AST) to all 100,000 selected clips.\footnote{Checkpoint: \url{https://huggingface.co/MIT/ast-finetuned-audioset-10-10-0.4593}.} Audio is decoded to mono at 16 kHz and processed in windows of at most 10 seconds, with a 5-second stride and an end-aligned final window. A clip receives a category tag when the maximum sigmoid score over its windows and the category's member classes is at least 0.5. Speech comprises the AudioSet Speech class and its descendants; music comprises Music, Singing, and their descendants. Other sounds comprise the remaining human, animal, natural, object, source-ambiguous, and noise events, excluding Silence and recording-context labels. The tags may overlap, and all percentages use 100,000 clips as the denominator.

These are automatic tagger-positive fractions. At the primary threshold, 18.973\% of clips remain unassigned. The estimates are sensitive to the decision threshold: at 0.3, the speech, music, and other-sound fractions are 65.046\%, 68.165\%, 13.061\%, respectively; at 0.7, they are 23.232\%, 21.159\%, 1.694\%, respectively.

\paragraph{Automatic shot counting.}

The shot-count summaries report automatic estimates of the number of visual segments per clip. A clip with $n$ internal shot boundaries contains $n+1$ shots. These counts describe visual segmentation; an audio event may continue across multiple shots.

\subsection{Temporal, Visual, and Audio Distributions}
\label{app:dataset_distributions}

The 5th and 95th percentiles of clip duration are 7.840 and 11.595 seconds. The automatic shot-count summaries contain 534,571 detected shots, and the 5th and 95th percentiles of shots per clip are 3 and 10. These statistics describe both clip length and the number of visual segments within each clip.

Table~\ref{tab:app_resolution_distribution} summarizes the distribution by the shorter image dimension: 51.832\% of clips fall in the 720--1,079-pixel bin, and 48.168\% have a short side of at least 1,080 pixels. 


The audio categories are reported as overlapping clip-level tags. At the primary threshold of 0.5, 25.672\% of clips receive both speech and music tags; this indicates detection of both categories somewhere within a clip, rather than necessarily at the same instant. The 18.973\% of clips receiving none of the three tags should not be interpreted as silent clips.

\subsection{Comparison with FineVideo~\citep{Farre2024FineVideo}}
Compared with FineVideo~\citep{Farre2024FineVideo},
MSAV-Script-100K exhibits stronger visual quality, audio quality,
and audio-visual alignment under the reported automatic metrics, as shown in Figure \ref{fig:msav_script_100k_info_appendix}.
For visual quality, it achieves higher mean LAION aesthetic
(5.471 vs.\ 4.349) and MUSIQ scores (66.344 vs.\ 56.517),
along with higher AMT-based motion smoothness
(0.9855 vs.\ 0.9789).
For audio quality, AudioBox assigns higher mean scores for
content enjoyment (5.335 vs.\ 4.655), content usefulness
(6.049 vs.\ 5.025), and production quality (6.920 vs.\ 5.869).
The mean Synchformer zero-offset probability is also higher
(0.301 vs.\ 0.266), indicating stronger estimated temporal
alignment between the visual and audio streams.
These advantages are reflected in both the means and medians
of all seven metrics, supporting the quality of
MSAV-Script-100K as a curated resource for multi-shot
audio-visual generation.

\begin{figure}[t]
    \centering
    \includegraphics[width=\linewidth]{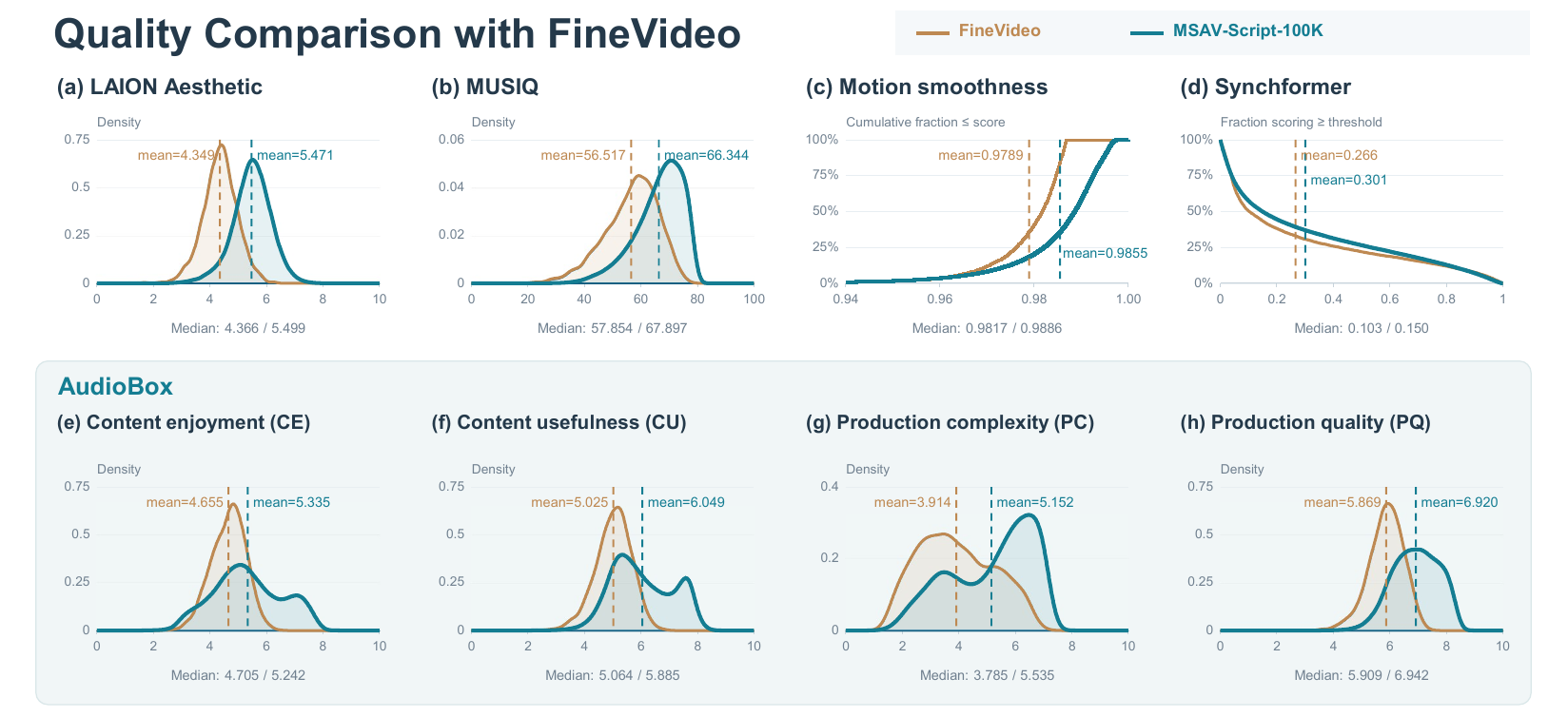}
    \vspace{-0.3 cm}
    \caption{Quality comparison between MSAV-Script-100K and FineVideo.}
    \vspace{-0.3 cm}
    \label{fig:msav_script_100k_info_appendix}
\end{figure}

\section{Filtering and Final Selection Details}
\label{app:dataset_filtering}

The curation procedure combines perceptual screening, semantic screening, and multi-criterion quality assessment. The following sections distinguish metric-based rejection from the later ranking of retained candidates.

\subsection{Resolution and Visual Screening}

\label{app:dataset_screening}

\paragraph{Resolution and visual screening.}
We retain clips whose shorter image dimension is at least 720 pixels, treating landscape and portrait orientations consistently. Table~\ref{tab:app_visual_screening} summarizes the subsequent six-metric visual filter. Let $P_q(m)$ denote the $q$th percentile of metric $m$ in the jointly valid reference population used to calibrate the visual filter. Calibration uses 242,109 clips with finite, valid values for all six metrics. Thresholds are computed once on this population and remain fixed through the successive filtering steps; equality at a threshold is retained. A clip must satisfy every applicable retention condition in the table. Percentile thresholds identify the tails of each metric distribution; because the rejected sets can overlap, the overall rejection rate is not the sum of the per-metric tail fractions.

\begin{table}[htbp]
\centering
\small
\caption{Visual screening criteria. Numerical thresholds are rounded for display; filtering uses the stored full-precision values. MUSIQ scores are divided by 100, and RAFT magnitudes are normalized by the evaluation-frame diagonal.}
\label{tab:app_visual_screening}
\begin{tabularx}{\linewidth}{@{}Xll@{}}
\toprule
Metric $m$ & Retention condition & Numerical threshold \\
\midrule
Sharpness & $m \geq P_5(m)$ & 10.278426 \\
Image entropy & $m \geq P_5(m)$ & 5.359885 \\
Flat-region proportion & $m \leq P_{95}(m)$ & 0.906250 \\
Texture score & $m \leq P_{95}(m)$ & 1.128293 \\
MUSIQ image quality & $m \geq P_5(m)$ & 0.391477 \\
RAFT motion magnitude & $P_5(m) \leq m \leq P_{95}(m)$ & $[0.004140,0.046509]$ \\
\bottomrule
\end{tabularx}
\end{table}

\paragraph{Spatial descriptors and sampling.}
Spatial metrics use frames sampled at 1 FPS, with uniform subsampling if more than 64 frames are available. Handcrafted descriptors are computed at the decoded spatial dimensions and averaged over frames. Sharpness is the variance of the grayscale Laplacian. Image entropy is the base-2 Shannon entropy of the 256-bin grayscale histogram. Flat-region proportion is the fraction of complete, non-overlapping $240\times240$ patches whose Sobel-gradient-magnitude variance is below 800; incomplete border patches are excluded. Texture is gray-level co-occurrence-matrix contrast after quantization to 16 levels, averaged over distance-one offsets at $0$, $\pi/4$, $\pi/2$, and $3\pi/4$, using symmetric, normalized matrices.
\paragraph{Learned visual metrics.}
MUSIQ uses the SPAQ checkpoint. Frames with a long side above 512 pixels are resized to that limit, and the clip score is the mean frame prediction divided by 100. RAFT uses \texttt{raft-things.pth} with 20 iterations on adjacent sampled frames at 8 FPS, capped at 256 frames per clip, with the long side limited to 512 pixels. Flow magnitude is averaged over pixels, divided by the resized image diagonal, and then averaged over frame pairs. The motion calculation includes pairs spanning shot transitions, so its upper-tail filter responds to editing cuts as well as within-shot motion.

\subsection{Low-Level Audio Screening}
\label{app:dataset_audio_screening}

We compute four descriptors to characterize audio before learned quality assessment: silence ratio, clipping ratio, an amplitude-based signal-to-noise proxy, and effective bandwidth. These descriptors measure waveform and spectral properties; speech quality and audio aesthetics are evaluated separately in Appendix~\ref{app:dataset_quality}.

The scorer operates on mono floating-point waveforms at the sampling rate of its evaluation input. The 16 kHz mono evaluation proxies have an 8 kHz Nyquist limit, so bandwidth measured on these proxies describes the evaluation signal. Analysis uses 2,048-sample frames with a 512-sample hop; inputs shorter than one frame are zero-padded. All clips entering this stage contain non-silent audio; the silence ratio measures relatively quiet intervals within each clip. For samples $y_n$ and frame RMS values $R_t$, the descriptors are:
\begin{itemize}
    \item \textbf{Silence ratio:} the fraction of frames satisfying $20\log_{10}(\max(R_t,10^{-12})/R_{\max})<-40$, where $R_{\max}=\max_tR_t$ and $R_t$ includes a $10^{-20}$ stabilizer inside the square root. This relative-amplitude measure is distinct from voice-activity detection.
    \item \textbf{Clipping ratio:} the fraction of samples with $|y_n|\geq0.99$.
    \item \textbf{Amplitude SNR proxy:} $10\log_{10}(P_{\mathrm{high}}/P_{\mathrm{low}})$, where the two powers are the mean squared amplitudes of samples at or above the 85th percentile and at or below the 15th percentile of $|y_n|$, respectively. The implementation returns zero when $P_{\mathrm{high}}=0$, and 100 dB when $P_{\mathrm{high}}>0$ and $P_{\mathrm{low}}\leq10^{-20}$. This statistic does not separate clean-signal power from noise power.
    \item \textbf{Effective bandwidth:} the first frequency bin at which cumulative power reaches 99\% of the time-averaged, one-sided power spectrum, using a Hann window. A zero-energy spectrum yields zero bandwidth.
\end{itemize}

We use these four audio descriptors to filter out clips with excessive silence, substantial waveform clipping, low amplitude contrast, or overly narrow effective bandwidth. This screening removes clips with undesirable waveform and spectral characteristics before the subsequent learned audio-quality assessment.

\subsection{Semantic Screening}
\label{app:dataset_semantic_screening}
We use Qwen3.5-27B~\citep{qwen2026qwen35} to assign a semantic meaningfulness score $s_{\mathrm{sem}}(x)\in[0,100]$ to each candidate clip $x$. We retain clips satisfying
\begin{equation}
    s_{\mathrm{sem}}(x) \geq 90.
    \label{eq:app_semantic_threshold}
\end{equation}
This assessment uses video frames only. It provides a content-level criterion beyond low-level visual properties, but does not directly verify speech intelligibility, sound-source correspondence, or audio continuity. These audio-related properties require the complementary audio and audio-visual assessments below.

\paragraph{Scoring criterion and visual inputs.}
The prompt evaluates visual semantic meaningfulness and editorial coherence. It rewards shots that form a coherent scene, action, event, narrative progression, montage, reaction/cutaway structure, or thematic sequence, and penalizes accidental concatenation or disconnected source fragments. Normal cuts, changes of viewpoint, flashbacks, and intentional discontinuities are acceptable when a plausible narrative, spatial, temporal, causal, thematic, or visual relationship remains. The 90--100 band denotes a clearly meaningful, coherent, and purposeful sequence. The prompt excludes blur, resolution, exposure, visual aesthetics, acting quality, and audio quality from this semantic judgment.

For scoring with ordered image inputs, we divide a clip of duration $D$ into eight equal-duration intervals and sample one frame at the center of each interval:
\begin{equation}
    t_i = \frac{i+1/2}{8}D, \qquad i=0,\ldots,7.
\end{equation}
The sampled frames are resized while preserving their aspect ratios, with the shorter side capped at 480 pixels. The eight frames are supplied to Qwen3.5-27B in chronological order together with their approximate timestamps, without audio. The model jointly assesses the sequence for meaningful relationships among shots, complementing the preceding frame-level visual quality measurement.

\paragraph{Score output.}
Inference uses temperature zero, and the model is instructed to return a single integer from 0 to 100. This score is compared directly with the retention threshold above.

\subsection{Tri-Dimensional Quality Assessment}
\label{app:dataset_quality}

After screening, we assess retained clips along three quality axes using the five criteria in Table~\ref{tab:app_quality_criteria}. The visual axis contributes two criteria (LAION aesthetics and AMT motion smoothness), the audio axis contributes two (AudioBox and DNSMOS/NISQA), and audio-visual synchronization contributes one (Synchformer).

\begin{table}[htbp]
\centering
\small
\caption{Five criteria used for quality assessment. Each criterion contributes one fifth of the overall score after within-source percentile conversion and criterion-level aggregation.}
\label{tab:app_quality_criteria}
\begin{tabularx}{\linewidth}{@{}p{0.18\linewidth}p{0.33\linewidth}X@{}}
\toprule
Axis & Criterion & Evaluator \\
\midrule
Visual & Visual aesthetics & LAION Aesthetic Predictor \\
Visual & Motion smoothness & AMT \\
Audio & Audio aesthetics and production quality & AudioBox \\
Audio & Speech quality & DNSMOS and NISQA \\
Audio-visual & Temporal synchronization & Synchformer zero-offset probability \\
\bottomrule
\end{tabularx}
\end{table}

\paragraph{Evaluator inputs and clip-level aggregation.}
Visual aesthetics uses CLIP ViT-L/14 with the LAION head \texttt{sa\_0\_4\_vit\_l\_14\_linear.pth}, averaging predictions from the spatial frame samples and dividing by ten. Motion smoothness uses AMT-S with \texttt{amt-s.pth}. Up to 32 uniformly spaced valid center indices form consecutive-frame triplets $(I_{t-1},I_t,I_{t+1})$. AMT predicts the middle frame from its neighbors; the clip score is one minus the mean absolute reconstruction error divided by 255. This is a sampled approximation rather than an evaluation of every frame.

AudioBox uses \texttt{facebook/audiobox-aesthetics} on 16 kHz mono audio in consecutive, non-overlapping 10-second windows. The final short window is zero-padded with a valid-sample mask. Predictions are restored to their native scales and averaged with weights proportional to valid window duration. Its four outputs are Content Enjoyment (CE), Content Usefulness (CU), Production Complexity (PC), and Production Quality (PQ). DNSMOS uses the calibrated OVRL output from \texttt{sig\_bak\_ovr.onnx}, with 16 kHz audio and 9.01-second windows at a one-second stride; short inputs are repeated until a complete window is available. NISQA uses \texttt{nisqa.tar}, its checkpoint-defined spectrogram configuration, and a 48 kHz evaluation waveform to produce MOS. Resampling an evaluation proxy does not restore frequency content absent from that proxy.

Synchformer uses checkpoint \texttt{24-01-04T16-39-21}, with 25 FPS video, 16 kHz audio, and zero imposed temporal offset. Its raw synchronization score is the softmax probability of the offset class nearest zero. This probability is converted to a percentile before entering the five-criterion average.

\paragraph{Within-source normalization and aggregation.}
Here, source groups are defined by the
original dataset or collection-category label in order to keep data diversity.
Raw evaluator outputs have different scales. For metric $m$, let $n_{s,m}$ be the number of finite, applicable values among candidates with source label $s$, and let $r_{s,m}(x)$ be the ascending rank of clip $x$, using average ranks for ties. We normalize each metric within the source label stored in the candidate metadata:
\begin{equation}
    p_{s,m}(x)=
    \begin{cases}
        (r_{s,m}(x)-1)/(n_{s,m}-1), & n_{s,m}>1,\\
        0.5, & n_{s,m}=1.
    \end{cases}
    \label{eq:app_metric_percentile}
\end{equation}
Suppressing the source index, the five subscores are
\begin{equation}
\begin{aligned}
    c_1(x)&=p_{\mathrm{LAION}}(x),\\
    c_2(x)&=p_{\mathrm{AMT}}(x),\\
    c_3(x)&=\frac{p_{\mathrm{CE}}(x)+p_{\mathrm{CU}}(x)+p_{\mathrm{PC}}(x)+p_{\mathrm{PQ}}(x)}{4},\\
    c_4(x)&=\frac{p_{\mathrm{DNSMOS\ OVRL}}(x)+p_{\mathrm{NISQA\ MOS}}(x)}{2},\\
    c_5(x)&=p_{\mathrm{sync}}(x).
\end{aligned}
    \label{eq:app_criterion_scores}
\end{equation}
The four AudioBox axes enter with the same positive sign. Their percentile average is our selection criterion, not an official AudioBox aggregate. The overall quality score is
\begin{equation}
    Q(x) = \frac{1}{5}\sum_{j=1}^{5} c_j(x).
    \label{eq:app_dataset_quality}
\end{equation}
Thus, equal weighting applies to the five criteria. The total weights of the visual, audio, and synchronization axes are $2/5$, $2/5$, and $1/5$, respectively. Within-source normalization measures a clip's relative standing within its source and reduces the influence of differences in raw score distributions across sources. It does not establish that clips with identical percentiles from different sources have identical absolute quality.

\paragraph{Scope of the quality signals.}
Each evaluator provides a complementary signal. Speech-quality scores are relevant when speech is present, whereas music, ambience, and sound effects also occur in clips without speech. Synchformer's zero-offset probability is used as a temporal-alignment signal; it does not by itself establish the correctness of every sound-source attribution or cross-shot event relation. Criterion-level scores retain the contributions of these distinct signals alongside the aggregate quality score.

\paragraph{Metric applicability.}
Speech evaluators run when the VAD record is valid and contains at least one detected speech segment with at least 0.5 seconds of total speech. Samples skipped by this gate receive a neutral percentile of 0.5 for each speech-quality metric, giving $c_4=0.5$. Synchformer requires at least five seconds of audio; an explicitly recorded short-input skip likewise receives a neutral synchronization percentile of 0.5. These assignments keep five terms in $Q(x)$ instead of changing their weights from clip to clip. The evaluator records distinguish completed predictions, applicability skips, and processing errors; an applicability skip is not a low-quality prediction.

\subsection{Final Selection}
\label{app:dataset_selection}

After visual and semantic screening and audio assessment, we select
100,000 clips through quality-based ranking within source groups,
allocating quotas proportionally to their candidate counts to preserve
source diversity. Within each group, clips are ranked in descending
order of the aggregate quality score $Q(x)$ defined in
Eq.~\ref{eq:app_dataset_quality}, with ties broken by ascending clip
identifier. Table~\ref{tab:app_source_composition} reports the source
composition of the final dataset.

\section{Qualitative Comparison and Analysis}
\label{app:qualitative}

We provide representative qualitative comparisons to complement the
quantitative evaluation. Figures~\ref{fig:msav_qualitative_comparison_case1}--\ref{fig:msav_qualitative_comparison_case5}
compare SSRS SFT with representative generation systems on multi-shot
event progression, while Figures~\ref{fig:SSRS_vs_Dense_example1}
--\ref{fig:SSRS_vs_Dense_example5} compare Dense SFT and SSRS SFT
under matched training conditions. Frames are sampled at the requested
shot windows; auditory relations in the controlled cases are additionally
evaluated using the native generated audio. \textbf{All qualitative examples are provided in the supplementary material.}

\paragraph{Ordered event progression.}
Figure~\ref{fig:msav_qualitative_comparison_case1} shows a four-stage
fire-lighting sequence: inspecting kindling, placing twigs, producing a
flame, and warming beside the fire. SSRS produces a clearer progression
from preparation to ignition and the resulting fire state, while competing
methods either repeat similar states, introduce the fire prematurely, or
drift from the requested event sequence. This case illustrates the benefit
of explicitly organizing local state changes through a global event order.

\paragraph{Multi-stage state evolution.}
Figure~\ref{fig:msav_qualitative_comparison_case2} considers planting
mint through five successive stages. SSRS more clearly assigns different
stages to different shots, progressing from preparation to planting,
soil treatment, and the final relocation toward light. In contrast,
other methods frequently repeat similar views or reveal later-stage
content too early. The comparison highlights that SSRS encourages
shot-specific states to evolve according to the intended temporal
structure rather than merely maintaining topical consistency.

\paragraph{Dialogue-driven shot organization.}
Figure~\ref{fig:msav_qualitative_comparison_case3} evaluates a museum
conversation involving walking together, a request from the man, and a
subsequent response from the woman. SSRS preserves the recurring
characters while adapting the shot composition to the evolving dialogue,
moving from a shared view to speaker-focused and response-focused views.
Other methods tend to repeat similar compositions or provide weaker
alignment between narrative roles and shot changes. This demonstrates
that cross-shot consistency should preserve identity while still allowing
the intended presentation and interaction states to change.

\paragraph{Role-consistent event progression.}
Figure~\ref{fig:msav_qualitative_comparison_case4} evaluates whether
persistent entity roles are maintained while actions evolve across
shots. The requested sequence requires the leader to point, the group
to run, and the same leader to finally grasp the flagpole. SSRS better
preserves the leader--group roles while following the intended action
order, whereas competing methods tend to repeat visually similar
states or provide weaker correspondence between entity roles and
requested actions. This example illustrates the benefit of jointly
organizing persistent identities and cross-shot event evolution.

\paragraph{Fine-grained object-state evolution.}
Figure~\ref{fig:msav_qualitative_comparison_case5} considers a
two-stage manipulation sequence in which a toolbox is first lowered
onto a cart and then secured by tightening its strap. SSRS maintains
the toolbox--cart relation while making the strap interaction and the
resulting secured state explicit. Although the pretrained LTX-2.3
already captures much of the requested sequence, SSRS more completely
realizes the fine-grained object states and interactions, demonstrating
improved adherence beyond coarse task completion.

\paragraph{Sound-source binding.}
Figure~\ref{fig:SSRS_vs_Dense_example1} isolates the relation between
a visual action and its designated sound source. The prompt requires only
the right blue bell to be struck in the second shot. SSRS realizes the
visible interaction with the specified bell and better preserves the bell
configuration across shots, whereas Dense SFT misses the requested strike
and later introduces an additional bell. Together with the native-audio
evaluation, this case illustrates the benefit of explicitly representing
source attribution and audio-visual temporal relations.

\paragraph{Cross-shot object-state persistence.}
Figure~\ref{fig:SSRS_vs_Dense_example2} tests whether a yellow backpack
remains associated with the green station bench across shot transitions.
Dense SFT instead places the backpack on a traveler, changing the intended
object--location relation. SSRS preserves the backpack--bench state in
the visible wide shots despite an intervening close-up in which the
backpack is not observable. This example highlights that temporary
visibility changes should not be interpreted as state changes, consistent
with SSRS's explicit separation of persistent identity from shot-local
visibility and state.

\paragraph{Non-speech sound-source binding.}
Figure~\ref{fig:SSRS_vs_Dense_example3} further evaluates
source-specific audio-visual grounding. The request requires the
right red bag to be unzipped in the second shot with a synchronized
zipper sound. SSRS realizes the interaction with the designated
red bag while preserving the surrounding scene, whereas Dense SFT
leaves both bags largely static. This example shows that SSRS can
better associate a non-speech auditory event with its intended
visual entity and action. The synchronization of the zipper sound
is verified using the native generated audio.

\paragraph{Speaker identity binding.}
Figure~\ref{fig:SSRS_vs_Dense_example4} examines whether speech is
assigned to the intended recurring character. The request specifies
that repairer B, identified by the blue workwear, speaks in the
second shot while repairer A remains silent. SSRS better preserves
the target character across shots and associates the requested
speech with repairer B, whereas Dense SFT exhibits substantial
identity drift in the later shot. This case illustrates the role of
shared visual and voice identities together with shot-level source
relations in maintaining speaker attribution. Speaker activity is
verified using the native generated audio.

\paragraph{Cross-shot audio continuity.}
Figure~\ref{fig:SSRS_vs_Dense_example5} evaluates whether an ongoing
auditory event persists across visual cuts. The request requires a
single acoustic-guitar passage to continue through two scene changes.
SSRS produces distinct visual views while maintaining the continuous
guitar event, whereas Dense SFT remains in nearly the same view
before failing to realize the final requested shot. This behavior
supports a central motivation of SSRS: visual shot boundaries need
not coincide with auditory event boundaries. Audio continuity is
verified using the native generated audio.

Overall, these examples show that the gains of SSRS are not limited to
visual similarity across shots. By explicitly representing persistent
entities, evolving states, audio-visual relations, and cross-shot event
organization, SSRS better follows both intended continuity and intended
change in multi-shot generation.
\begin{figure}[t]
    \centering
    \includegraphics[width=\linewidth]{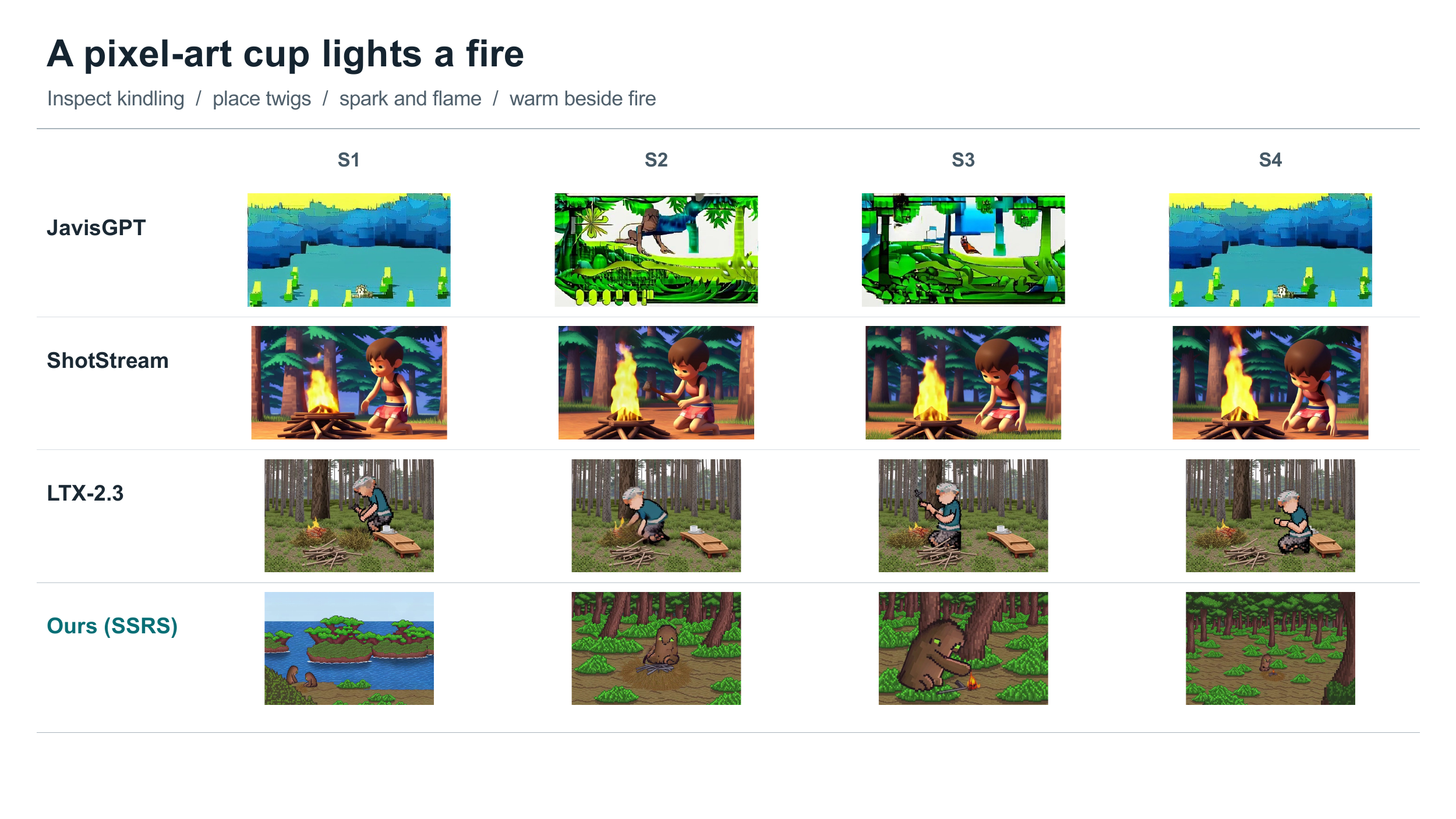}
    \vspace{-1.5cm}
    \caption{\textbf{Ordered event progression.}
    SSRS better follows the intended fire-lighting sequence, while
    competing methods exhibit repeated, premature, or weakly ordered states.}
    \vspace{-0.2cm}
    \label{fig:msav_qualitative_comparison_case1}
\end{figure}
\begin{figure}[t]
    \centering
    \includegraphics[width=\linewidth]{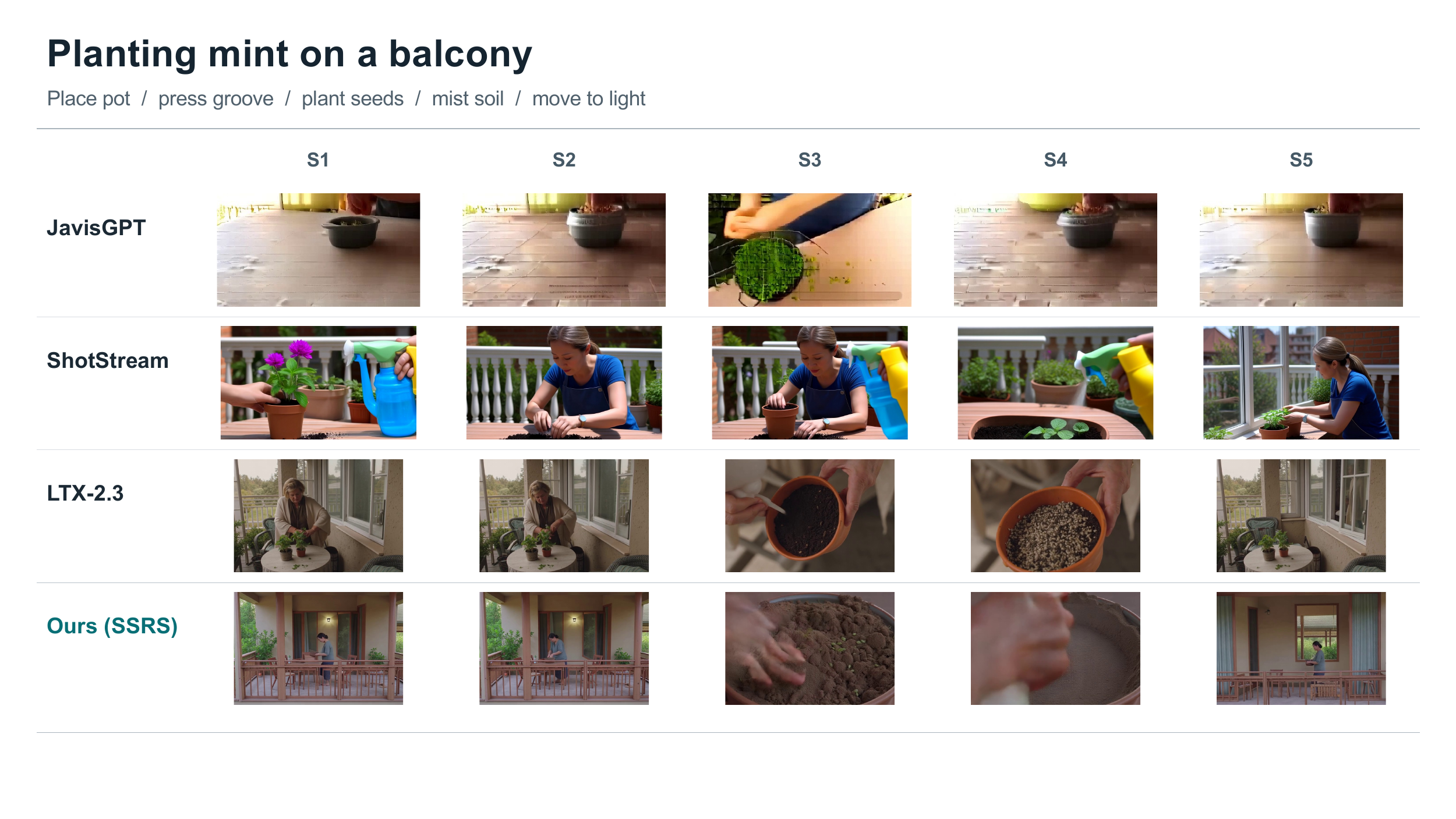}
    \vspace{-1.5cm}
    \caption{\textbf{Multi-stage state evolution.}
    SSRS more clearly aligns successive planting states with the requested
    shot sequence, whereas competing methods frequently repeat or collapse stages.}
    \vspace{-0.2cm}
    \label{fig:msav_qualitative_comparison_case2}
\end{figure}
\begin{figure}[t]
    \centering
    \includegraphics[width=\linewidth]{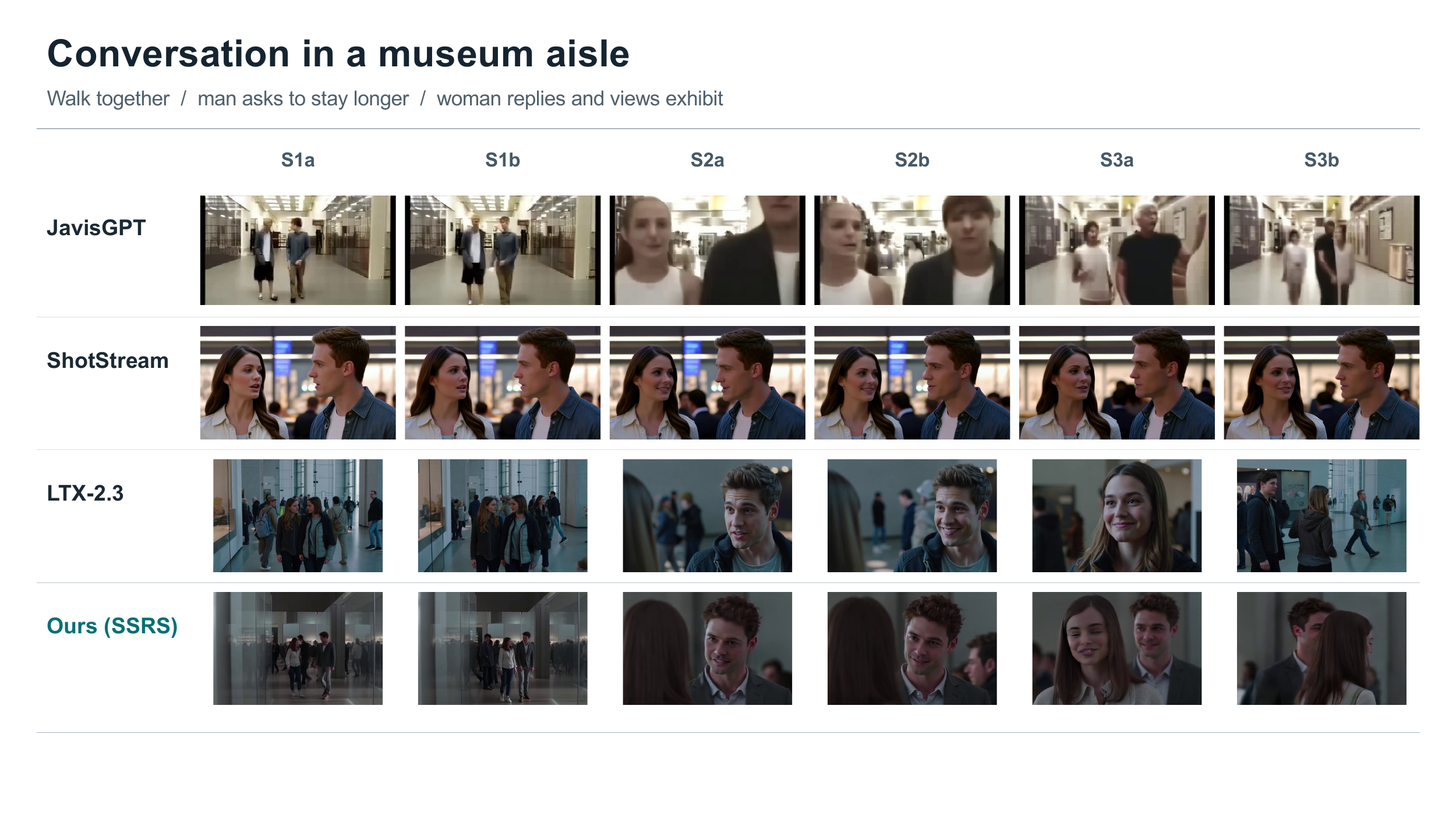}
    \vspace{-1.5cm}
    \caption{\textbf{Dialogue-driven shot organization.}
    SSRS preserves recurring identities while adapting shot composition
    to changing speaker and interaction states.}
    \vspace{-0.2cm}
    \label{fig:msav_qualitative_comparison_case3}
\end{figure}

\begin{figure}[t]
    \centering
    \includegraphics[width=\linewidth]{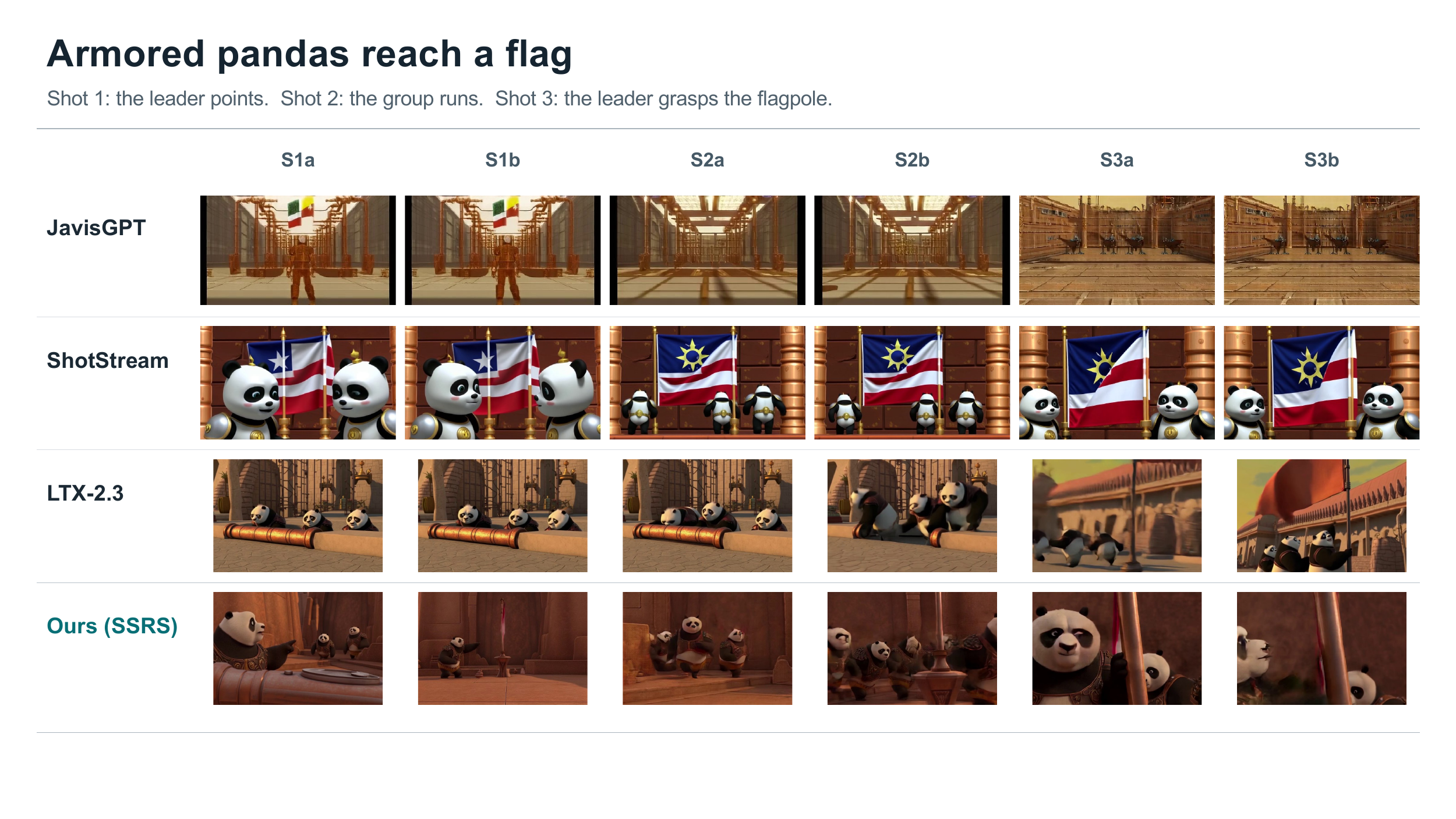}
    \vspace{-1.5cm}
    \caption{\textbf{Role-consistent cross-shot event progression.}
    SSRS preserves the leader identity while following the requested
    point--run--grasp sequence across shots.}
    \vspace{-0.2cm}
    \label{fig:msav_qualitative_comparison_case4}
\end{figure}

\begin{figure}[t]
    \centering
    \includegraphics[width=\linewidth]{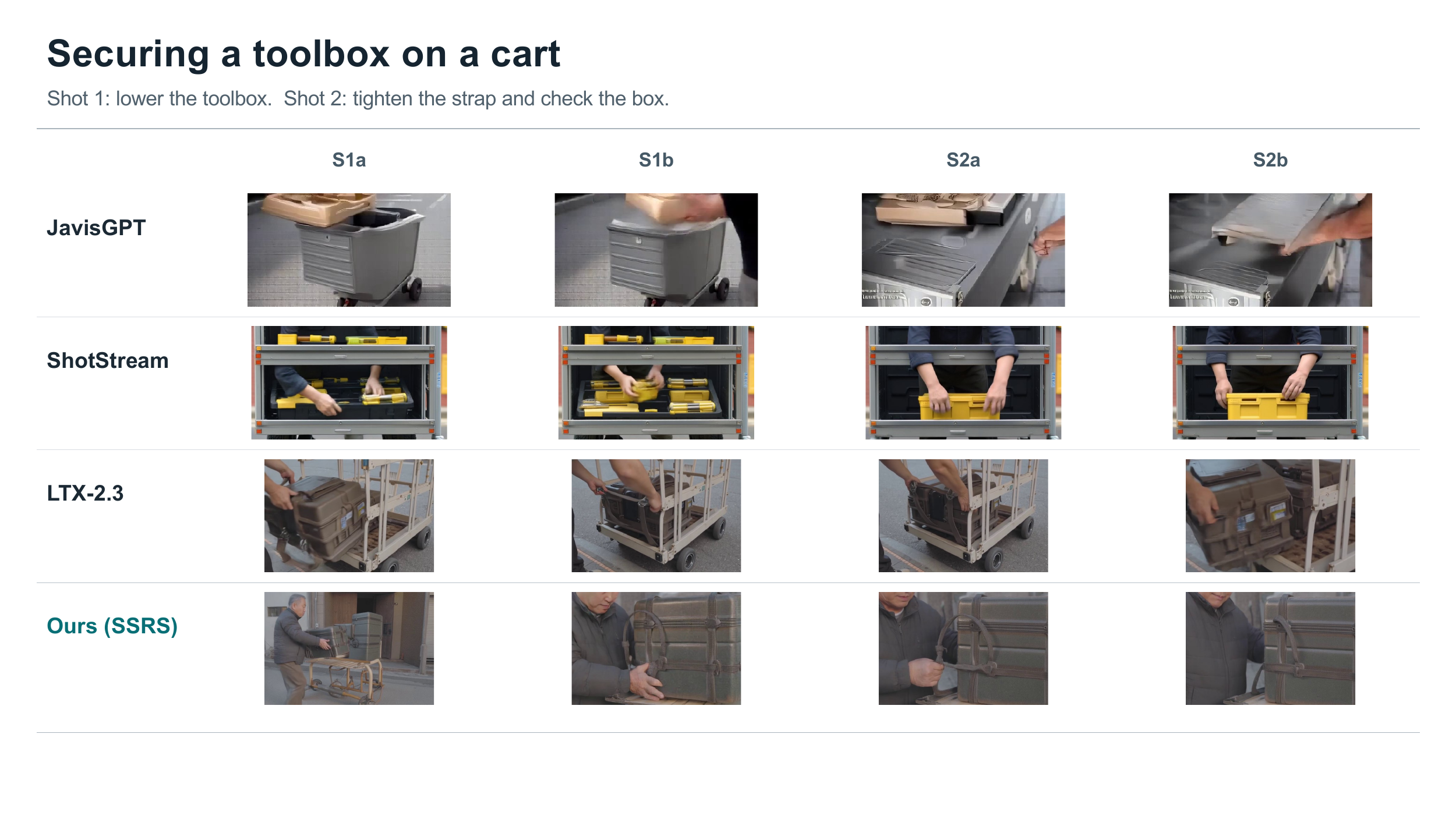}
    \vspace{-1.5cm}
    \caption{\textbf{Fine-grained object-state evolution.}
    SSRS maintains the toolbox--cart relation while explicitly realizing
    the requested lowering, strap-tightening, and secured final state.}
    \vspace{-0.2cm}
    \label{fig:msav_qualitative_comparison_case5}
\end{figure}

\begin{figure}[t]
    \centering
    \includegraphics[width=\linewidth]{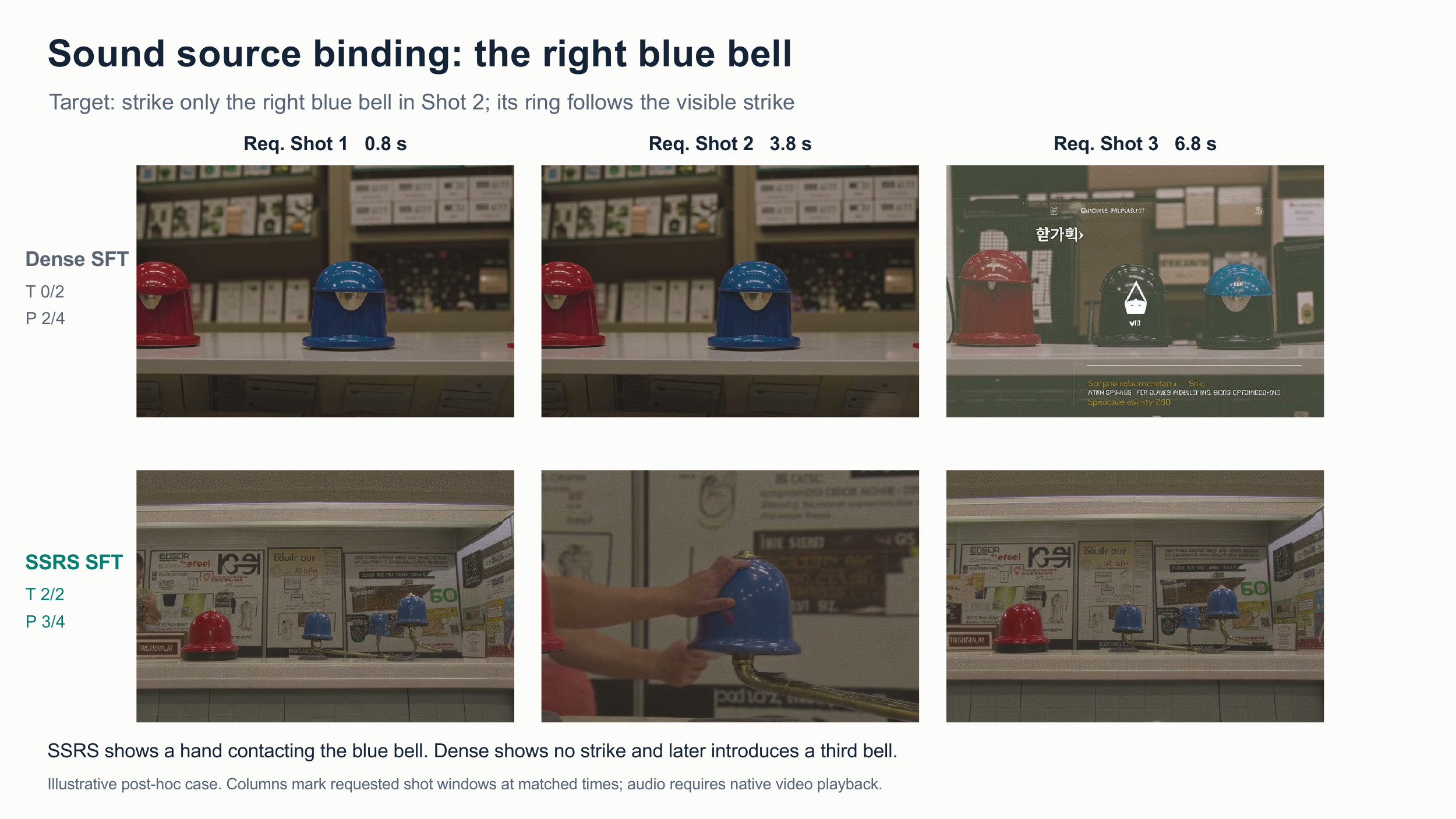}
    \vspace{-0.9cm}
    \caption{\textbf{Sound-source binding.}
    SSRS realizes the requested interaction with the designated blue bell,
    whereas Dense SFT misses the strike and exhibits object drift across shots.}
    \vspace{-0.2cm}
    \label{fig:SSRS_vs_Dense_example1}
\end{figure}
\begin{figure}[t]
    \centering
    \includegraphics[width=\linewidth]{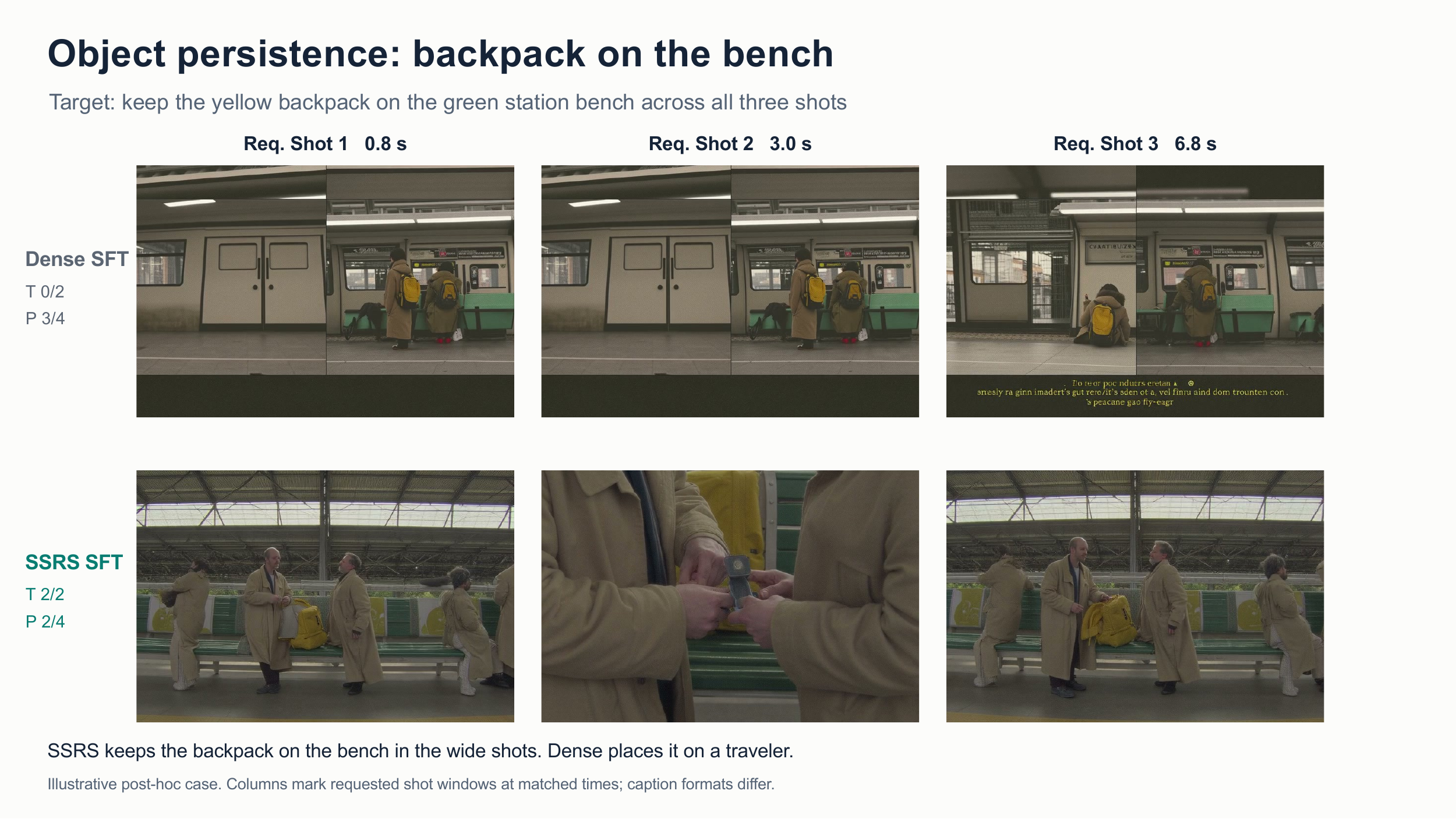}
    \vspace{-0.9cm}
    \caption{\textbf{Cross-shot object-state persistence.}
    SSRS preserves the backpack--bench relation across shot transitions,
    whereas Dense SFT incorrectly transfers the backpack to a traveler.}
    \vspace{-0.2cm}
    \label{fig:SSRS_vs_Dense_example2}
\end{figure}

\begin{figure}[t]
    \centering
    \includegraphics[width=\linewidth]{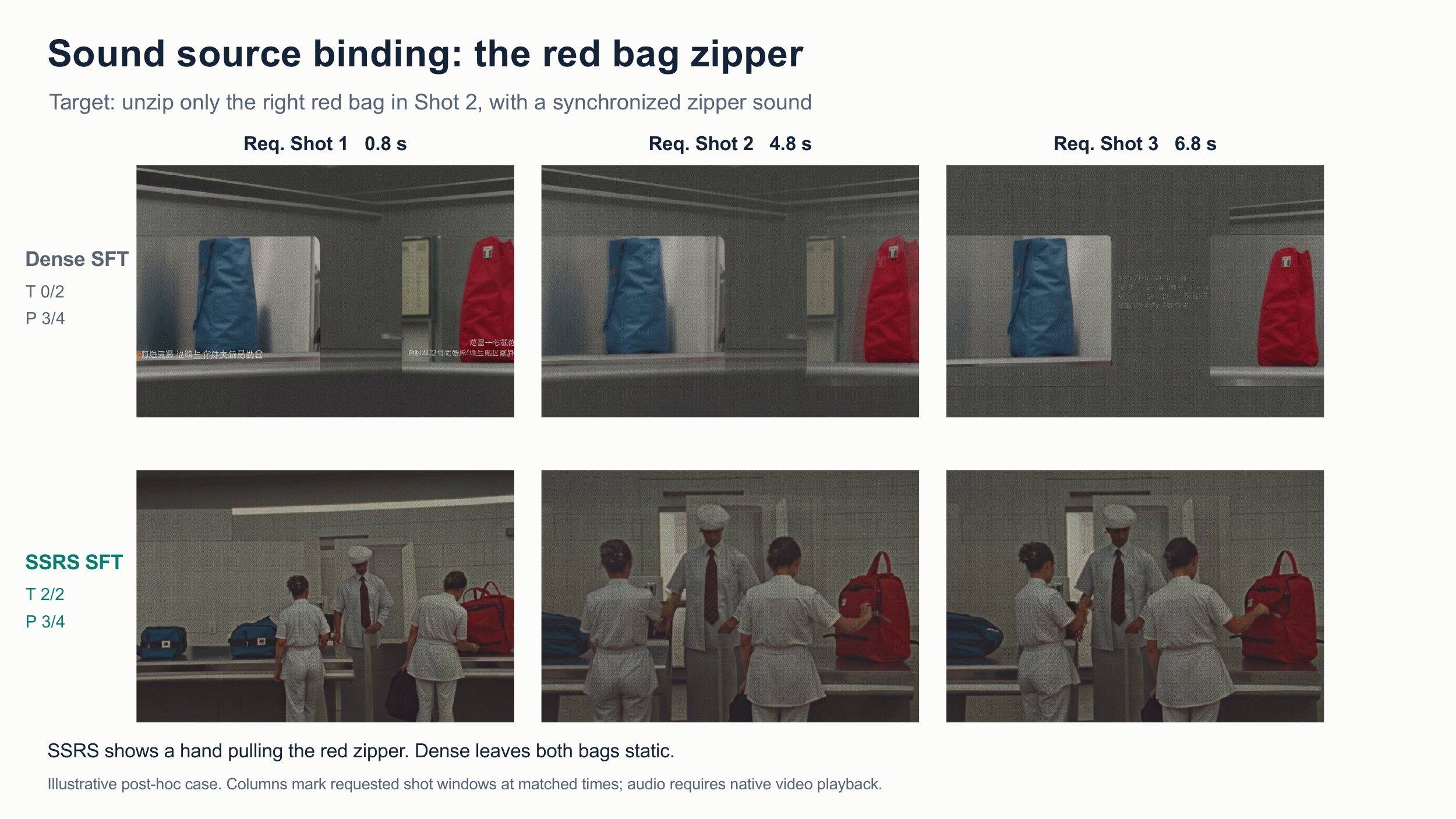}
    \vspace{-0.9cm}
    \caption{\textbf{Non-speech sound-source binding.}
    SSRS realizes the requested zipper interaction on the designated
    red bag, whereas Dense SFT leaves both bags largely static.
    Zipper synchronization is verified using the native generated audio.}
    \vspace{-0.2cm}
    \label{fig:SSRS_vs_Dense_example3}
\end{figure}
\begin{figure}[t]
    \centering
    \includegraphics[width=\linewidth]{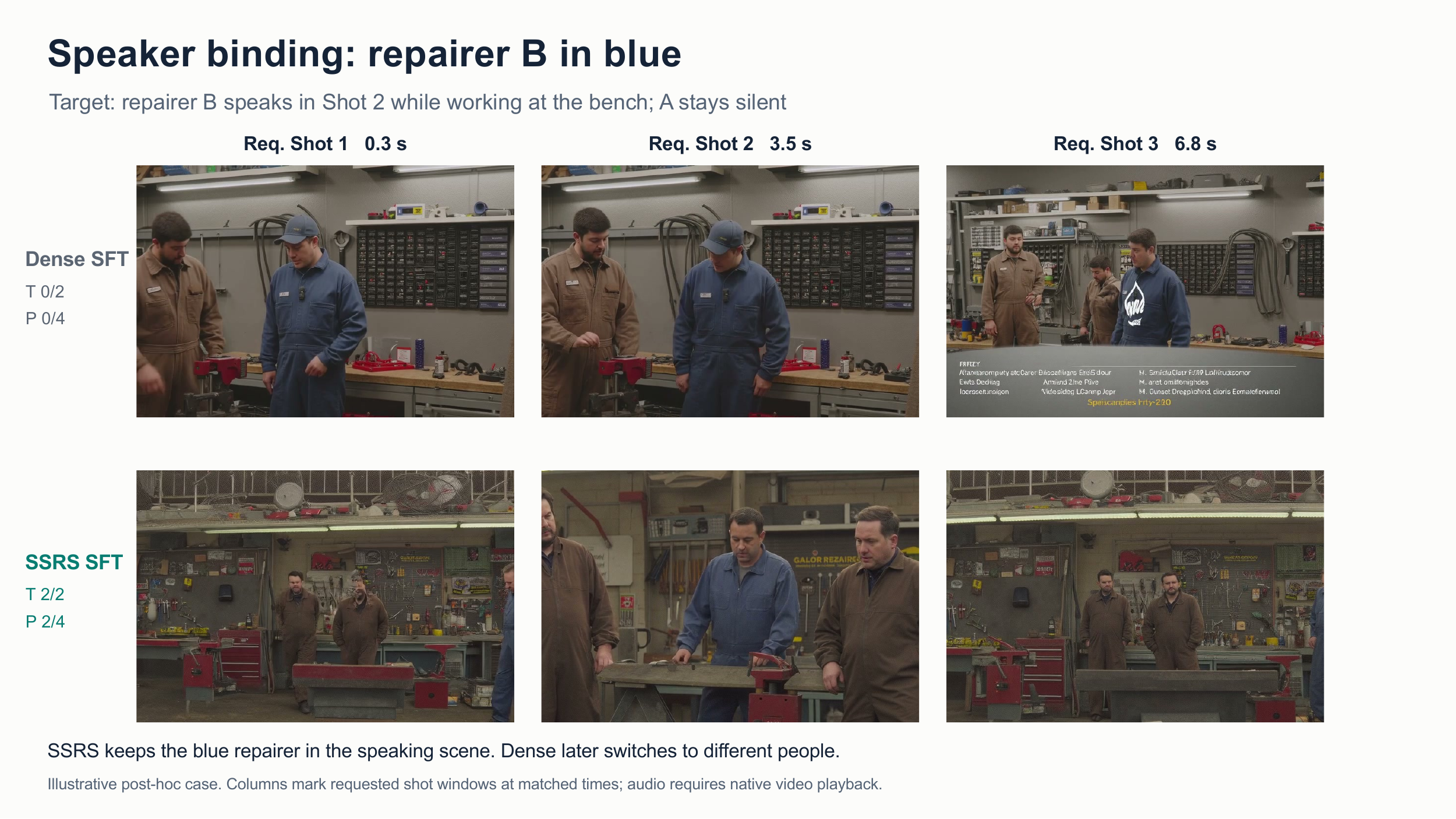}
    \vspace{-0.9cm}
    \caption{\textbf{Speaker identity binding.}
    SSRS preserves the target repairer across shots and associates
    the requested speech with repairer B, whereas Dense SFT exhibits
    identity drift. Speaker activity is verified using the native audio.}
    \vspace{-0.2cm}
    \label{fig:SSRS_vs_Dense_example4}
\end{figure}

\begin{figure}[t]
    \centering
    \includegraphics[width=\linewidth]{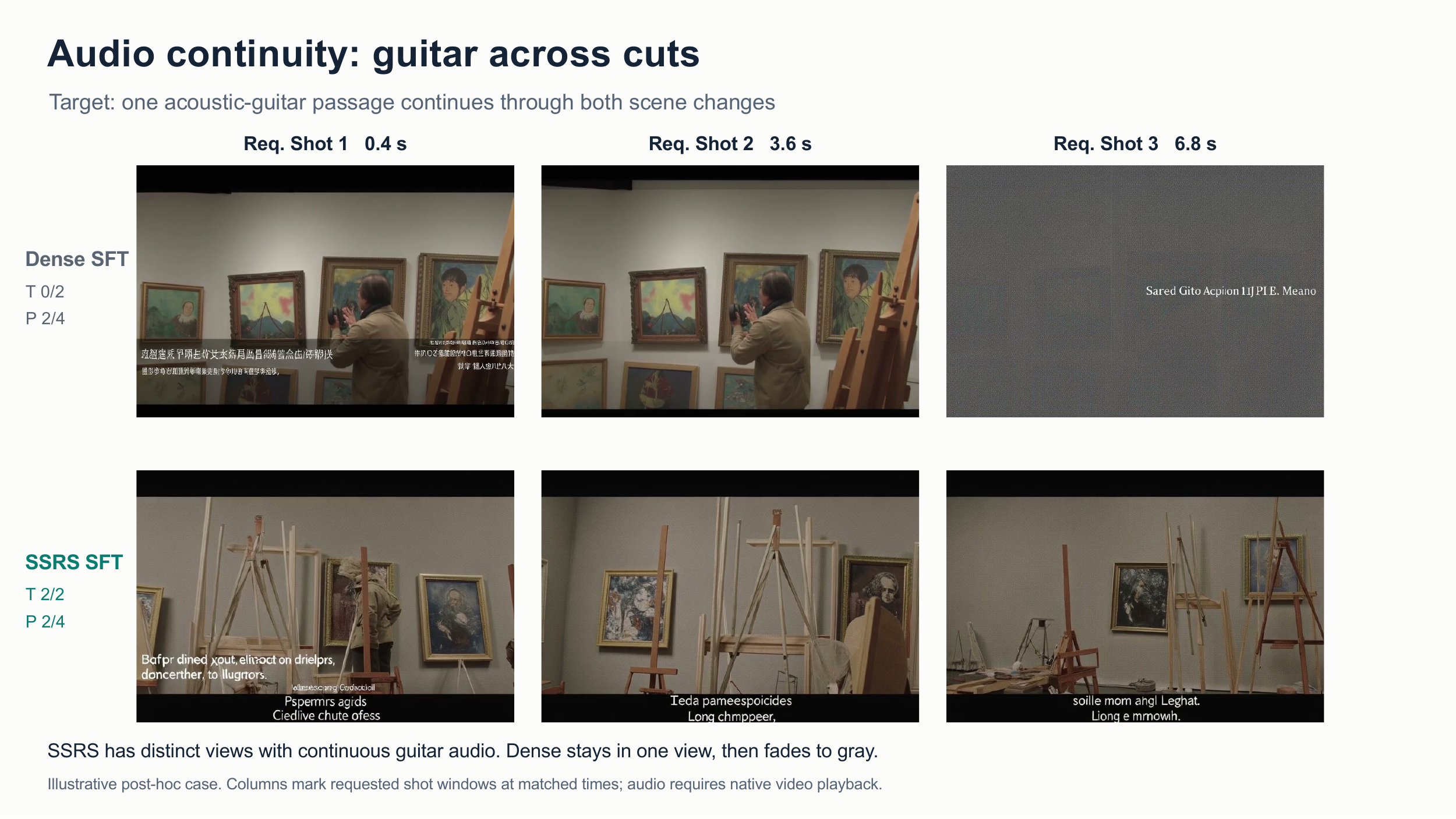}
    \vspace{-0.9cm}
    \caption{\textbf{Cross-shot audio continuity.}
    SSRS maintains a single guitar event across changing visual views,
    illustrating that auditory events can persist across shot boundaries,
    whereas Dense SFT fails to realize the requested multi-shot progression.}
    \vspace{-0.2cm}
    \label{fig:SSRS_vs_Dense_example5}
\end{figure}

\section{Details of the Controlled State and Relation Evaluation}
\label{app:controlled_eval}

\paragraph{Diagnostic design.}
The controlled evaluation is designed to directly probe the
stateful and relational behaviors targeted by SSRS, rather than
serve as an additional general-purpose benchmark. We evaluate
40 matched prompt instances, producing 40 Dense SFT outputs and
40 SSRS SFT outputs. The instances cover five diagnostic
categories: visual-state evolution (VS), cross-shot
audio continuity (AC), non-speech sound-source binding
(NS), cross-shot event order (EO), and
on-/off-screen speaker binding (SB). Because the category
sizes are small and unbalanced, category-level results are treated
as descriptive diagnostic evidence rather than statistically
well-powered benchmark estimates.

\paragraph{Questions and scoring.}
For each generated video, we predefine two \emph{target questions}
and four \emph{preservation questions}. Target questions test
whether the requested state transition or relation is realized,
whereas preservation questions test whether non-target requirements
remain satisfied. Gemini~3.5~Flash observes the generated video
together with its native audio and answers the fixed binary
questions. Requested content that is absent from the generation
is counted as unsatisfied rather than inapplicable.

For video $i$, we define
\begin{equation}
T_i = \frac{1}{2}\sum_{q=1}^{2}\mathbf{1}[y^{T}_{iq}=\mathrm{yes}],
\qquad
P_i = \frac{1}{4}\sum_{q=1}^{4}\mathbf{1}[y^{P}_{iq}=\mathrm{yes}],
\end{equation}
where $y^{T}_{iq}$ and $y^{P}_{iq}$ denote the answers to target
and preservation questions, respectively. Category-level and
overall scores are obtained by averaging $T_i$ and $P_i$ over
the corresponding matched prompt instances. Dense SFT and SSRS
SFT are evaluated using the same underlying requirements and
question sets.

\paragraph{Diagnostic dimensions.}
VS evaluates whether an entity follows a requested state change
while preserving its identity and other shared attributes.
AC tests whether an auditory event correctly continues or changes
across visual cuts. NS evaluates whether a non-speech sound is
assigned to the intended visual source. EO tests the specified
ordering of events across shots. SB evaluates whether speech is
correctly attributed to an on-screen or off-screen speaker,
including cases in which the visible listener differs from the
continuing speaker.

\paragraph{Additional analysis.}
SSRS improves target satisfaction in every category:
VS by 40.00 pp, AC by 50.00 pp, NS by 40.00 pp, EO by
40.00 pp, and SB by 35.71 pp. Preservation also improves for
AC, NS, EO, and SB. VS shows the opposite preservation trend:
$T$ increases from 20.00\% to 60.00\%, whereas $P$ decreases
from 62.50\% to 42.50\%. This suggests that stronger compliance
with requested visual-state changes may occasionally perturb
non-target visual properties. The overall preservation rate,
however, remains comparable and slightly higher for SSRS
(52.50\% vs.\ 49.38\%).

These results should be interpreted as a controlled diagnostic
analysis. In particular, the relatively small numbers of EO
and SB instances limit fine-grained statistical conclusions.
The primary evidence is therefore the consistent increase in
target satisfaction across all five independently defined
state and relation categories.

\section{Training Details}
\label{sec:Training}

\paragraph{Multi-Shot Audio-Visual Captioner.}
Table~\ref{tab:captioner_hyperparameters} summarizes the key
hyperparameters used to fine-tune our Multi-Shot Audio-Visual
Captioner. The captioner is initialized from the Qwen3-Omni backbone
and adapted using LoRA for parameter-efficient fine-tuning. Training is
conducted on 8 NVIDIA A100 GPUs with 80\,GB memory for three epochs,
taking approximately 68 hours. We use the AdamW optimizer with a peak
learning rate of $1.0\times10^{-4}$ and a cosine learning-rate schedule.
A warmup ratio of 0.05 is employed to stabilize the initial stage of
cross-modal alignment learning.

\begin{table}[h]
\centering
\caption{Training hyperparameters of the Multi-Shot Audio-Visual Captioner.}
\label{tab:captioner_hyperparameters}
\begin{tabular}{lc}
\toprule
\textbf{Hyperparameter} & \textbf{Configuration} \\
\midrule
Backbone                 & Qwen3-Omni             \\
GPU Usage                & $8\times$ A100 80\,GB  \\
Batch Size per GPU       & 1                      \\
Gradient Accumulation    & 2                      \\
Effective Batch Size     & 16                     \\
Training Time            & 68 hours               \\
Training Epochs          & 3                      \\
Peak Learning Rate       & $1.0\times10^{-4}$     \\
Learning Rate Scheduler  & Cosine                 \\
Optimizer                & AdamW                  \\
Fine-tuning Type         & LoRA                   \\
Warmup Ratio             & 0.05                   \\
\bottomrule
\end{tabular}
\end{table}

\begin{table}[h]
\centering
\caption{Training hyperparameters of the LTX-2.3 audio-visual DiT.}
\label{tab:dit_hyperparameters}
\begin{tabular}{lc}
\toprule
\textbf{Hyperparameter} & \textbf{Configuration} \\
\midrule
Backbone                 & LTX-2.3 22B            \\
GPU Usage                & $2\times8$ H800 80\,GB \\
Batch Size per GPU       & 1                      \\
Gradient Accumulation    & 1                      \\
Effective Batch Size     & 16                     \\
Training Resolution      & $768\times512$         \\
Frames per Sample        & 121                    \\
Training Frame Sampling  & Adaptive full-duration \\
Audio Processing         & Temporally aligned     \\
Maximum Training Steps   & 20,000                \\
Learning Rate            & $1.0\times10^{-4}$     \\
Learning Rate Scheduler  & Linear                 \\
Optimizer                & AdamW                  \\
Maximum Gradient Norm    & 1.0                    \\
Numerical Precision      & BF16                   \\
Gradient Checkpointing   & Enabled                \\
Fine-tuning Type         & LoRA                   \\
LoRA Rank                & 64                     \\
LoRA Alpha               & 64                     \\
LoRA Dropout             & 0.0                    \\
LoRA Target Modules      & Q/K/V/Output projections \\
Text Condition           & Precomputed            \\
Audio-Video Latents      & Precomputed            \\
\bottomrule
\end{tabular}
\end{table}
\paragraph{Audio-Visual Diffusion Transformer.}
We further fine-tune the LTX-2.3 22B audio-visual diffusion transformer
using the captions generated by our Multi-Shot Audio-Visual Captioner. Table~\ref{tab:dit_hyperparameters} summarizes the key
hyperparameters used to fine-tune our Audio-Visual Diffusion Transformer.
The DiT is adapted using LoRA, while the original model parameters
remain frozen. LoRA modules are inserted into the query, key, value,
and output projections of the attention layers. Both the LoRA rank and
scaling parameter are set to 64, and no LoRA dropout is applied\cite{zhao2026spiking,zhou2026learning,zhao2025zero,hu2025exploiting,zhao2024cycle,zhao2025learning,lu2025does,11513016,Zhaowfdiff,zhao2026learning,zhao2025ultrahr,zhao2026luve,zhang2026probing,zhangrethinking,zhang2025gapt,zhang2025r,zhang2024vocapter,zhang2025u}.

Training is distributed over two compute nodes, each equipped with
eight NVIDIA H800 GPUs with 80\,GB memory, resulting in a total of 16
GPUs. Each GPU processes one sample per optimization step, giving an
effective global batch size of 16 without gradient accumulation. We
use AdamW with a learning rate of $1.0\times10^{-4}$, a linear
learning-rate schedule, and a maximum gradient norm of 1.0. Training is
performed in BF16 precision with gradient checkpointing enabled.

Each training sample contains 121 video frames at a spatial resolution
of $768\times512$. Rather than enforcing a fixed frame rate during
training, we uniformly sample 121 frames over the complete duration of
each source clip. Therefore, the effective sampling frame rate is
adaptively determined by the source duration. The temporally aligned
audio segment is encoded over the same interval. This strategy retains
the complete temporal structure of clips with slightly different
durations while maintaining a fixed tensor shape for distributed
training.

To reduce repeated computation, text conditions and audio-video VAE
latents are encoded once and cached before DiT optimization. The
training stage consequently loads only the cached conditions and
latents, avoiding repeated execution of the text encoder and
audio-video VAE encoders. The cache is verified for completeness,
identity consistency, and finite tensor values before optimization.

\section{Preference Optimization of the  Captioner}
\label{app}

\paragraph{Preference-pair construction.}
We perform direct preference optimization (DPO) after supervised fine-tuning (SFT) of the Qwen3-Omni-based captioner. For each multi-shot clip in the preference-data subset, the SFT captioner samples 20 dense captions from the same audio-visual input. Sampling uses a temperature of 0.7, top-$p$ of 0.9, top-$k$ of 50, and a maximum of 2,048 generated tokens. We then provide Gemini~3.1~Pro with these candidates and the clip's expert-constructed dense caption as a textual reference. The judge is asked to identify a preferred caption and three dispreferred captions exhibiting, respectively, the most severe visual errors, audio errors, and audio-visual alignment or temporal-logic errors. This produces up to three preference pairs per clip, each sharing the same preferred response but targeting a different failure mode. Unlike the initial annotation pipeline, this preference-selection step compares captions against the reference text; the judge does not directly re-inspect the clip.

Across four multi-shot subsets, 3,781 clips yielded 11,341 pair records. Each record contains the clip's video and audio paths, the captioning instruction, a preferred response, and one dispreferred response. The preference data are used only to refine the captioner that produces dense descriptions; the subsequent dense-to-SSRS conversion is a separate stage.

\paragraph{Optimization details.}
We initialize DPO from the merged SFT checkpoint of Qwen3-Omni-30B-A3B and use that SFT model as the reference policy. Training uses the standard sigmoid DPO objective~\citep{rafailov2023dpo} with $\beta=0.1$. We apply LoRA to the language model with rank~8, scaling factor~16, and zero LoRA dropout, while freezing the vision tower and multimodal projector. Training runs for three epochs, or 534 optimizer steps, on 32 H800 GPUs across four nodes with DeepSpeed ZeRO-3 and bfloat16 precision. The per-device batch size is one preference pair, with two gradient-accumulation steps, giving an effective batch size of 64 pairs. We use a learning rate of $10^{-4}$, a cosine schedule, a 0.05 warm-up ratio, and a maximum sequence length of 22,000 tokens. The resulting captioner is used to annotate the larger clip collection before SSRS conversion.

\section{SSRS Conversion and Quality Control}
\label{sec:appendix_ssrs_conversion}

We convert dense audio-visual captions into SSRS through a
typed JSON intermediate representation, followed by deterministic
text rendering.
The conversion is designed to reorganize source information into
explicit entities, states, and relations while preserving the
source content and avoiding unsupported additions.
This section describes the representation, conversion procedure,
and automated quality-control mechanisms.

\paragraph{Canonical JSON representation.}
The intermediate representation contains global visual and audio
attributes, a visual entity registry, an audio identity registry,
an audio entity registry, an ordered shot sequence, and a global
event sequence.
Visual entities store their categories and appearance attributes.
Audio identities describe reusable voice characteristics, while
audio entities specify speech, sound effects, ambience, and music,
including their content and acoustic properties.
Transcribed speech entries preserve the source transcript, and
speech entries reference reusable voice identities when supported
by the source.
Speech without an available transcript is represented through
its transcription status rather than an invented utterance.
Non-linguistic vocalizations are distinguished through a subtype.

Each shot records available temporal boundaries, camera attributes,
frame details, visual and audio states, and within-shot relations.
The global event sequence connects these entries through temporal
and relational descriptions, including simultaneity, persistence,
cross-shot continuity, and source-supported triggering relations.
Registries provide reusable definitions, whereas shot-level fields
and the global event sequence describe their contextual states
and interactions.

The schema specifies objects, arrays, strings, enumerated values,
and nullable references.
Visual entities, audio identities, audio entities, shots, and
global events use separate identifier namespaces:
\texttt{v1}, \texttt{p1}, \texttt{a1}, \texttt{s1}, and
\texttt{e1}, respectively.
Identifiers must be unique within their namespaces, and references
must resolve to entries of the expected type.
The supported audio entity types are
\texttt{speech}, \texttt{sfx}, \texttt{ambience}, and \texttt{music}.
Source evidence and quality-control metadata are retained for
auditing but excluded from the rendered training caption.

\paragraph{Language-model conversion.}
We use \texttt{Qwen3.5-35B-A3B} to produce the structured JSON.
The conversion protocol comprises system instructions, an output
schema, an initial conversion prompt, and a feedback-based repair
prompt.
The source caption serves as the reference for conversion:
the model is instructed to preserve its stated facts, dialogue,
temporal information, negation, and uncertainty.
Registries define reusable entities, while shot-level fields
describe their local states and relationships.

Relations are included only when supported by the source.
The co-occurrence of an entity and a sound does not by itself
establish a sound-source binding, and temporal succession does
not by itself establish a triggering relation.
The conversion instructions require unspecified information
to remain unspecified through optional or nullable fields,
and require uncertainty expressed in the source to be retained.


\paragraph{Deterministic text rendering.}
A rule-based renderer traverses the JSON in a fixed order,
maps internal identifiers to readable labels, and formats the
global, registry, shot-level, and event-sequence fields into
SSRS text.
Field labels follow the selected language convention.
Rendering does not invoke another language model.
For a fixed renderer version and configuration, the same JSON
therefore produces the same textual representation.
This separates structured content generation from text formatting
and makes the rendering stage reproducible.

\paragraph{Automated quality control and repair.}
Quality control combines structural validation, source-string
consistency checks, and targeted rules for common conversion errors.
Structural checks cover JSON validity, schema compliance,
identifier uniqueness, reference integrity, and relation-endpoint
validity.
Source-consistency checks examine transcript correspondence and
agreement with unambiguous shot-count and temporal anchors
extracted from the source caption.
Additional rules flag selected audio-type inconsistencies and
violations of the separation between reusable registry definitions
and shot-level relationships.
Evidence strings are checked against the source text and associated
with source locations when possible.

Validation findings provide feedback for the repair stage.
The repair prompt directs the model to address the reported
violations while preserving source-supported content.
Repaired outputs undergo the same validation checks.


\subsection{Representative Dense-to-SSRS Examples}

The following examples are compact, human-verified pairs.
They preserve the facts in the displayed dense captions while
omitting implementation-only metadata and evidence spans for readability.

\subsubsection{Example 1}

\paragraph{Dense caption.}
The video contains two shots. From 0.0--1.0 s, a static, eye-level
medium shot shows a boy at center stage in a green grasshopper
costume, with a girl in a leaf costume dimly visible to his right.
Warm stage lighting illuminates a white fence, a prop tree, and a
shimmering curtain. The boy looks forward and says, ``I'm a
grasshopper,'' with visible lip synchronization. The recording is
dry and close, with little reverberation. From 1.0--6.0 s, a static,
eye-level medium close-up shows a couple in the audience against
softly blurred spectators. A man in a dark blue shirt looks toward
the stage and says, ``I told you.'' He turns to the smiling woman in
a floral dress and continues, ``He is the cutest one.'' The woman
gives a soft verbal response whose words are not provided. The man
then looks back toward the stage while the woman continues looking
forward. The lighting and overall atmosphere are warm, intimate,
and affectionate, and the close, dry recording continues.

\paragraph{SSRS caption.}
\textbf{Global Visual.}
Scene and space: a stage area and an audience area.
Lighting and atmosphere: warm lighting is present across the
sequence; the audience shot has a warm, intimate, and affectionate
atmosphere.

\textbf{Global Audio.}
Overall soundscape: close, dry dialogue recording with little
reverberation.
Global auditory atmosphere: not specified.
Non-diegetic music: not specified.

\textbf{Visual Entity Registry.}
[Subject A] A boy wearing a green grasshopper costume.
[Subject B] A girl wearing a leaf costume.
[Subject C] A man wearing a dark blue shirt.
[Subject D] A woman wearing a floral dress.
[Object A] A white fence.
[Object B] A prop tree.
[Object C] A shimmering curtain.

\textbf{Audio Identity Registry.}
[Voiceprint A] Voice identity used by [Voice 1].
[Voiceprint B] Shared voice identity used by [Voice 2] and [Voice 3].
[Voiceprint C] Voice identity used by [Voice 4].

\textbf{Audio Entity Registry.}
[Voice 1] uses [Voiceprint A]; transcript: ``I'm a grasshopper.''
[Voice 2] uses [Voiceprint B]; transcript: ``I told you.''
[Voice 3] uses [Voiceprint B]; transcript: ``He is the cutest one.''
[Voice 4] uses [Voiceprint C]; transcript: unavailable;
a soft verbal response.

\textbf{Shot 1 (0.0--1.0 s).}
Presentation: static, eye-level medium shot with warm stage lighting.
Frame details: [Subject A] is at center stage; [Subject B] is dimly
visible to his right. [Object A], [Object B], and [Object C] are
present on the stage.
Visual state and change: [Subject A] looks forward and speaks.
Audio state: [Voice 1] occurs.
Relations: [Subject A] is the visible source of [Voice 1], and the
visible lip motion of [Subject A] is synchronized with [Voice 1].

\textbf{Shot 2 (1.0--6.0 s).}
Presentation: static, eye-level medium close-up with warm lighting;
spectators in the background are softly blurred.
Frame details: [Subject C] and [Subject D] are shown in the audience.
Visual state and change: [Subject C] looks toward the stage and says
[Voice 2], turns toward [Subject D] and says [Voice 3], and then
looks back toward the stage. [Subject D] smiles, gives [Voice 4],
and continues looking forward.
Audio state: [Voice 2] precedes [Voice 3], followed by [Voice 4].
Relations: [Subject C] is the visible source of [Voice 2] and
[Voice 3]. [Subject D] is the visible source of [Voice 4].

\textbf{Global Event Sequence.}
E1: [Shot 1] [Subject A] says [Voice 1] on stage.
$\rightarrow$
E2: [Shot 2] [Subject C] says [Voice 2] and [Voice 3], after which
[Subject D] gives [Voice 4].
The visual setting changes from the stage to the audience, while
the close, dry dialogue recording continues.

\subsubsection{Example 2}

\paragraph{Dense caption.}
The video contains four shots in a bright, rustic outdoor cooking
setting. From 0.0--1.0 s, a static, slightly low-angle close-up
shows a left hand holding a green bell pepper on a wooden cutting
board while a right hand slices it with a metal chef's knife; the
rhythmic blade impacts are clearly audible. From 1.0--4.8 s, a
static, eye-level medium shot reveals a man in a dark
camouflage-patterned jacket behind a table of vegetables, bowls,
and a terracotta jug. He continues slicing the green pepper as the
chopping sound continues, accompanied by a low wind rumble.
From 4.8--6.0 s, a static close-up shows a basket containing red
pepper rings and halves while the wind continues. From 6.0--8.0 s,
pale yellow pepper slices are added to the same basket, producing
a crisp rustling sound as they settle; the wind remains audible.
Bright natural sunlight and an earthy, tranquil atmosphere persist
throughout.

\paragraph{SSRS caption.}
\textbf{Global Visual.}
Scene and space: a rustic outdoor cooking setting containing a
cutting board, a table of vegetables and bowls, a terracotta jug,
and a basket.
Lighting and atmosphere: bright natural sunlight and an earthy,
tranquil atmosphere persist throughout.

\textbf{Global Audio.}
Overall soundscape: rhythmic blade impacts, a low wind rumble,
and the crisp rustling of pepper slices settling.
Global auditory atmosphere: not specified.
Non-diegetic music: not specified.

\textbf{Visual Entity Registry.}
[Subject A] A man wearing a dark camouflage-patterned jacket.
[Object A] A green bell pepper.
[Object B] A metal chef's knife.
[Object C] A wooden cutting board.
[Object D] A table containing vegetables, bowls, and a terracotta jug.
[Object E] A basket containing red pepper rings and halves.
[Object F] Pale yellow pepper slices.

\textbf{Audio Identity Registry.}
None.

\textbf{Audio Entity Registry.}
[SFX 1] Rhythmic blade impacts associated with slicing the green
pepper on the cutting board.
[Ambience 1] A low wind rumble.
[SFX 2] Crisp rustling produced as the pale yellow pepper slices
settle in the basket.

\textbf{Shot 1 (0.0--1.0 s).}
Presentation: static, slightly low-angle close-up in bright natural
sunlight.
Visual state and change: a left hand holds [Object A] on [Object C]
while a right hand repeatedly slices it with [Object B].
Audio state: [SFX 1] occurs repeatedly.
Relations: [SFX 1] is associated with the repeated slicing action
involving [Object B], [Object A], and [Object C].

\textbf{Shot 2 (1.0--4.8 s).}
Presentation: static, eye-level medium shot.
Frame details: [Subject A] is behind [Object D].
Visual state and change: [Subject A] continues slicing [Object A]
with [Object B] on [Object C].
Audio state: [SFX 1] continues; [Ambience 1] is present.
Relations: [SFX 1] remains associated with the ongoing slicing
action; [Ambience 1] is present in the outdoor scene.

\textbf{Shot 3 (4.8--6.0 s).}
Presentation: static close-up.
Visual state: [Object E] contains red pepper rings and halves.
Audio state: [Ambience 1] continues.
Relations: [Ambience 1] remains present in the outdoor scene.

\textbf{Shot 4 (6.0--8.0 s).}
Presentation: not further specified in the source caption.
Visual state and change: [Object F] is added to [Object E] and
settles.
Audio state: [SFX 2] occurs while [Ambience 1] continues.
Relations: the addition and settling of [Object F] produces
[SFX 2]; [Ambience 1] remains audible.

\textbf{Global Event Sequence.}
E1: [Shot 1--Shot 2] [Object A] is sliced, with [SFX 1] continuing
across the cut.
$\rightarrow$
E2: [Shot 3] [Object E] is shown while [Ambience 1] continues.
$\rightarrow$
E3: [Shot 4] [Object F] is added to [Object E], producing [SFX 2].
[Ambience 1] persists from [Shot 2] through [Shot 4].

\begin{table*}[t]
\centering
\caption{Definitions and computation of the 14 metrics in our
MSAVBench-derived evaluation.
All normalized scores lie in $[0,1]$; higher is better.}
\label{tab:msavbench_metrics}

\scriptsize
\setlength{\tabcolsep}{3pt}
\renewcommand{\arraystretch}{1.10}

\begin{tabularx}{\textwidth}{
@{}l
>{\raggedright\arraybackslash}p{0.13\textwidth}
>{\raggedright\arraybackslash}X
>{\raggedright\arraybackslash}p{0.30\textwidth}@{}
}
\toprule
Metric & Dimension & Measurement & Normalization \\
\midrule

\multicolumn{4}{@{}l}{\textit{Global metrics}} \\
\addlinespace[2pt]

Narr
& Narrative coherence
& Binary visual judgments of requested events and their
narrative order.
& $B(\mathcal Q_{\mathrm{narr}})$. \\

\addlinespace[2pt]
Lip
& Lip--speech synchronization
& SyncNet synchronization confidence $c$ and signed
frame offset $o$ for matched speaking tracks.
& $0.7C(c/6)+0.3C(1-|o|/8)$;
aggregated across tracks using visual-duration weights. \\

\addlinespace[2pt]
Attr
& Speaker attribution
& Match diarized audio speakers to active visual speaker tracks.
& $\sum_k d_k\mathbb{I}[m_k]/\sum_k d_k$,
where $d_k$ is speech duration and $m_k$ indicates
a successful assignment. \\

\addlinespace[2pt]
Sync
& Audiovisual synchronization
& Synchformer estimates signed delays
$\delta_{\mathrm{first}}$ and $\delta_{\mathrm{last}}$
for the first and last feature blocks, in seconds.
& $C(1-d/2)$, where
$d=(|\delta_{\mathrm{first}}|
+|\delta_{\mathrm{last}}|)/2$. \\

\addlinespace[2pt]
VQ
& Visual realization
& Answer a shared set of visual multiple-choice questions
constructed from the evaluation prompt.
& Number of correct answers divided by the total
number of questions. \\

\midrule
\multicolumn{4}{@{}l}{\textit{Cross-shot metrics}} \\
\addlinespace[2pt]

C-Layout
& Cross-shot layout
& Tool-assisted binary judgments of spatial relationships
across aligned shots.
& $B(\mathcal Q_{\mathrm{cross}})$. \\

\addlinespace[2pt]
VC
& Visual consistency
& Compare subject appearance using localized DINOv2-L features,
background using subject-removed CLIP ViT-L/14 features,
and style using CSD-ViT-L features.
Assess illumination and color with Gemini binary judgments.
& Mean of five components:
$C((1+\bar c)/2)$ for subject and background,
$C(\bar c)$ for style, and
$B(\mathcal Q)$ for illumination and color. \\

\addlinespace[2pt]
Mus
& Music consistency
& Extract accompaniment with Demucs and compare
MuQ embeddings, tempo, and beat intervals across audio windows.
Tempo and beats are estimated with All-In-One.
& $C(0.55S_{\mathrm{emb}}
+0.25S_{\mathrm{bpm}}
+0.20S_{\mathrm{beat}})$;
components are defined below. \\

\addlinespace[2pt]
Spk
& Speaker consistency
& Extract W2V-BERT speaker embeddings and compare all unordered
window pairs belonging to the same diarized speaker.
& Map pairwise cosine similarity using
$C((1+\cos)/2)$, average within each speaker,
and aggregate speaker scores by speech duration. \\

\midrule
\multicolumn{4}{@{}l}{\textit{Local metrics}} \\
\addlinespace[2pt]

I-Layout
& Intra-shot layout
& Tool-assisted binary judgments of object positions
and spatial relationships within each shot.
& Average $B(\mathcal Q_{s,\mathrm{layout}})$
across evaluated shots. \\

\addlinespace[2pt]
Cam
& Camera adherence
& Binary judgments of adherence to requested framing,
viewpoint, and camera motion.
& Average $B(\mathcal Q_{s,\mathrm{cam}})$
across evaluated shots. \\

\addlinespace[2pt]
PQ
& Audio production quality
& Use the production-quality output $q$ of
\texttt{facebook/audiobox-aesthetics},
with a nominal rating scale of 1--10.
& $C((q-1)/9)$. \\

\addlinespace[2pt]
OCR
& Visual text fidelity
& Match OCR detections to requested text strings using
normalized-string similarity $r_t$ and the selected
detection's confidence $c_t$.
& $\frac{1}{T}\sum_{t=1}^{T}
r_t(0.5+0.5c_t)$,
where $T$ is the number of requested text targets. \\

\addlinespace[2pt]
WER-S
& Spoken-content fidelity
& Compare ASR output with requested transcripts using
word error rate, or character error rate for Chinese.
& $1-\min(E,1)$, where
$E=(N_{\mathrm{sub}}+N_{\mathrm{del}}
+N_{\mathrm{ins}})/N_{\mathrm{ref}}$. \\

\bottomrule
\end{tabularx}

\vspace{4pt}
\begin{minipage}{\textwidth}
\scriptsize

\textbf{Notation.}
$C(x)=\min(1,\max(0,x))$;
$B(\mathcal Q)=N_{\mathrm{yes}}/
(N_{\mathrm{yes}}+N_{\mathrm{no}})$.
$\bar c$ denotes mean cosine similarity.
For WER, $N_{\mathrm{sub}}$, $N_{\mathrm{del}}$,
$N_{\mathrm{ins}}$, and $N_{\mathrm{ref}}$ denote the numbers
of substitutions, deletions, insertions, and reference units,
respectively. The reported WER-column score is normalized
speech fidelity, rather than raw error rate.

\smallskip
\textbf{Music components.}
For retained music-window embeddings $\mathbf z_i$,
\[
\begin{aligned}
S_{\mathrm{emb}}
&=\frac{2}{n(n-1)}
  \sum_{i<j}
  C\!\left(\frac{1+\cos(\mathbf z_i,\mathbf z_j)}{2}\right),\\
S_{\mathrm{bpm}}
&=C\!\left(1-\frac{\sigma_b}{\max(1,\mu_b)}\right),
\qquad
S_{\mathrm{beat}}
=C\!\left(1-\frac{\sigma_\tau}{\mu_\tau+10^{-6}}\right).
\end{aligned}
\]
Here, $\mu_b$ and $\sigma_b$ are the population mean and
standard deviation of positive BPM estimates.
Each $\tau_i$ is the mean inter-beat interval of a window
containing at least two detected beats; $\mu_\tau$ and
$\sigma_\tau$ are the population mean and standard deviation
of these window-level means.
$S_{\mathrm{beat}}$ measures interval consistency,
rather than beat-phase alignment.

\smallskip
\textbf{Audio windows.}
Spk uses 4-s windows with a default stride of 4 s
and a maximum of eight windows, retaining speaker--window
chunks containing at least 1 s of speech.
For clips longer than 32 s, the eight window starts are
evenly distributed from $0$ to $D-4$, where $D$ is clip duration.
Mus uses 4-s windows with a maximum of four windows and
discards decoded trailing segments of at most 0.5 s.

\smallskip
\textbf{OCR matching.}
Text preprocessing applies NFKC normalization, lowercasing,
and punctuation/whitespace filtering.
String matching uses a substring-match rule with score 1
and otherwise computes
\texttt{difflib.SequenceMatcher.ratio()}.
The best string match determines $r_t$ and its associated
detection confidence $c_t$.

\end{minipage}
\end{table*}

\section{Evaluation Metrics and Implementation Details}
\label{sec:appendix_evaluation}

We evaluate multi-shot audio-visual generation using 14 metrics
adapted from MSAVBench.
Table~\ref{tab:msavbench_metrics} summarizes their measurement
procedures and normalization.
We describe the implementation-specific choices below to make
the resulting scores interpretable and reproducible.

\paragraph{Evaluation dimensions.}
The metrics cover three complementary levels.
Global metrics assess narrative coherence (Narr), lip--speech
synchronization (Lip), speaker--visual attribution (Attr),
general audiovisual synchronization (Sync), and realization of
requested visual attributes (VQ).
Cross-shot metrics assess spatial consistency (C-Layout),
visual consistency (VC), music consistency (Mus), and speaker
identity consistency (Spk).
Local metrics assess within-shot spatial relationships (I-Layout),
camera instruction adherence (Cam), perceptual audio quality (PQ),
on-screen text fidelity (OCR), and spoken-content fidelity (WER).
VQ measures visual attribute realization rather than aesthetic
quality. VC averages five components: subject appearance,
background, style, illumination, and color.

\paragraph{Visual judging and protocol adaptations.}
We use \texttt{gemini-3.5-flash} for visual judgments.
Direct visual evaluation uses an 8-fps silent video proxy with
a maximum width of 640 pixels, while expert audiovisual metrics
retain the original media.
VQ questions are constructed from the canonical evaluation prompt
before inspecting generated videos and are shared across models
evaluated on that prompt.
Shot boundaries are obtained from visual observations by the judge,
rather than the original benchmark's full shot-detection and
iterative-correction pipeline.
Music and speaker consistency use fixed audio windows, which
need not coincide with visual shot boundaries.

\paragraph{Score normalization.}
All normalized scores in
Table~\ref{tab:msavbench_metrics} lie in $[0,1]$, with higher
values indicating better performance.
Binary rubric scores average applicable yes/no judgments.
Expert outputs use metric-specific mappings rather than a shared
rescaling rule.
In particular, Lip combines synchronization confidence and frame
offset, whereas Sync penalizes delay measured in seconds.
OCR uses confidence-weighted string similarity.
For spoken-content fidelity, the normalized score is
$1-\min(E,1)$, where $E$ is WER, or CER for Chinese.
These mappings differ from some descriptions in the original
protocol; raw and normalized quantities should therefore not be
interchanged.

\end{document}